\documentclass{article}
\usepackage{iclr2027_conference,times}
\iclrfinalcopy
\usepackage[T1]{fontenc}
\usepackage{amsmath,amssymb,bm}
\usepackage{graphicx}
\usepackage{booktabs,colortbl,multirow,arydshln}
\usepackage{soul,pifont}
\usepackage[skins,listings]{tcolorbox}
\usepackage{capt-of,wrapfig}
\usepackage{algorithm,algpseudocode}
\usepackage{hyperref,url}
\usepackage{enumitem}
\definecolor{caseback}{HTML}{F8F9FB} \definecolor{caserule}{HTML}{4D4D4D} \definecolor{stageone}{HTML}{1F6FB4} \definecolor{stagetwo}{HTML}{D1622B} \definecolor{stagethree}{HTML}{2E8B57} \definecolor{caseflag}{HTML}{8A6BBE} \definecolor{hlclue}{HTML}{FFF0A6} \definecolor{hlans}{HTML}{BFE7C8}

\newcommand{\hlc}[1]{{\sethlcolor{hlclue}\hl{#1}}}
\newcommand{\hla}[1]{{\sethlcolor{hlans}\hl{#1}}}

\newtcolorbox{casebox}[1]{%
  enhanced, sharp corners=downhill, arc=1.5pt,
  colback=caseback, colframe=caserule, boxrule=0.6pt,
  left=4pt, right=4pt, top=3pt, bottom=3pt,
  fonttitle=\bfseries\scriptsize, coltitle=white, colbacktitle=caserule,
  title={#1}}

\newtcolorbox{casebar}[1]{%
  enhanced, sharp corners=downhill, arc=1.5pt,
  colback=caseback, colframe=caserule, boxrule=0.6pt,
  left=3pt, right=3pt, top=2pt, bottom=2pt,
  fonttitle=\bfseries\scriptsize, coltitle=white, colbacktitle=caserule,
  title={#1}}

\newcommand{\stage}[2]{\vspace{1.5pt}\noindent{\color{#1}\scriptsize\bfseries #2}\par\vspace{-1pt}}
\newcommand{\qa}[2]{\noindent{\scriptsize\textbf{Question.}~#1\\[1pt]\textbf{Ground-truth answer.}~#2}\par}

\newenvironment{tracelist}
  {\begin{list}{$\bullet$}{\leftmargin=1.1em \itemindent=0em \labelsep=0.35em
    \topsep=1pt \parsep=0pt \itemsep=0.5pt \scriptsize}}
  {\end{list}}

\lstdefinestyle{promptstyle}{%
  basicstyle=\rmfamily\footnotesize,
  breaklines=true, columns=fullflexible, keepspaces=true,
  showstringspaces=false, upquote=true,
  literate={—}{{\textemdash}}1}

\newtcblisting{promptbox}[2][stageone]{%
  enhanced, sharp corners,
  colback=white, colframe=caserule!40, boxrule=0.4pt,
  left=6pt, right=6pt, top=5pt, bottom=5pt,
  fonttitle=\bfseries\small, coltitle=#1, colbacktitle=caseback,
  title={#2}, listing only, listing options={style=promptstyle}}

\newtheorem{proposition}{Proposition}
\algrenewcommand{\algorithmiccomment}[1]{\hfill\textcolor{black!50}{$\triangleright$ #1}}
\algrenewcommand{\algorithmicrequire}{\textcolor{stageone}{\textbf{Require:}}}
\algrenewcommand{\algorithmicensure}{\textcolor{stageone}{\textbf{Return:}}}

\definecolor{ourrow}{gray}{0.93}
\definecolor{lossred}{RGB}{170,40,40} \definecolor{okgreen}{RGB}{30,120,60}

\title{Policy-Adaptive Credit Assignment for Spatial Reasoning with Verifiable Rewards}

\author{Hailiang Yang\\
School of Computer Science and Engineering\\
University of New South Wales\\
\texttt{z5516532@ad.unsw.edu.au}
\And
Shengchao Chen\thanks{Corresponding author and project leader.}\\
Australian AI Institute\\
University of Technology Sydney\\
\texttt{shengchao.chen.uts@gmail.com}
}
\hypersetup{pdfauthor={Hailiang Yang and Shengchao Chen}}

\begin{document}
\maketitle
\fancyhead{}
\renewcommand{\headrulewidth}{0pt}

\begin{abstract}
Image-based spatial reasoning requires perception, decomposition, and inference, but the stage that limits accuracy can change during training. Answer-level rewards neither localize failures to individual stages nor provide discriminative group-relative feedback when all rollouts receive the same reward. Process supervision provides intermediate feedback, but fixed weights cannot adapt reward allocation to these changing bottlenecks. To address this gap, we propose CALIPER, an adaptive process-reward framework supported by SpatialTrace, a corpus of stage-decomposed spatial reasoning traces. CALIPER jointly estimates the associations between stage-wise reference coverage and answer correctness through within-group ridge regression, using these associations as utility proxies to gradually update reward weights. The resulting process reward supervises all rollout groups, including uniformly failing ones. Because utility estimates can become unreliable when answer-discriminative groups are scarce, a closed-loop harness monitors estimator reliability and validation performance to guide checkpoint rollback, rollout diversification, and early stopping. Extensive experiments across eight benchmarks and three backbones show that CALIPER achieves the highest mean under both one-shot and harness training, with consistent gains over fixed allocation across RL optimizers. Further analyses confirm the benefits of adaptive reward allocation and harness control, with favorable scaling in model/data size.
\end{abstract}

\section{Introduction}
Image-based spatial reasoning remains a persistent weakness of vision-language models, spanning metric and relational reasoning in 3D scenes \citep{ma2024threedsrbench}, open-domain scene understanding \citep{zhang2024mmerealworld} and region-level search in high-resolution images \citep{wu2024v}. Although typically evaluated as a direct mapping from an image and question to an answer, solving these tasks often involves distinct stages of perception, decomposition and inference, whose relative importance varies across instances. Reinforcement learning from verifiable rewards (RLVR) \citep{guo2025deepseekr1}, often using group-relative advantages over multiple rollouts of the same question \citep{shao2024deepseekmath}, has been extended to multimodal reasoning by Vision-R1 \citep{huang2026vision} and DeepEyes \citep{zheng2026deepeyes}. However, its answer-level supervision has two limitations. It cannot distinguish failures in perception, decomposition or inference, and it provides no discriminative signal when all rollouts receive the same reward, especially on difficult spatial instances.

\begin{figure}[tbh]\centering
\includegraphics[width=.97\textwidth]{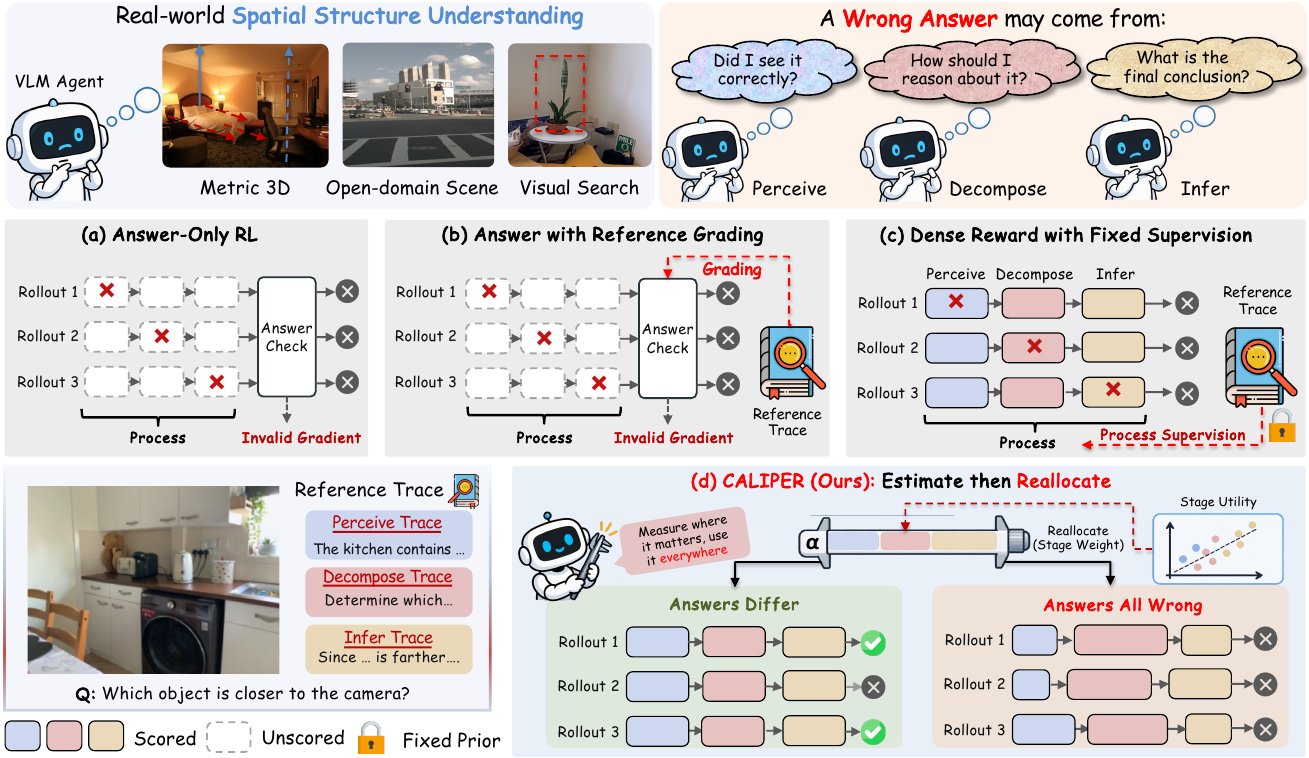}%
\vspace{-8pt}
\caption{\small Outcome-only rewards (a), as in GRPO \citep{shao2024deepseekmath} and DeepSeek-R1 \citep{guo2025deepseekr1}, supervise final answers. Reference-assisted verification (b) improves grading but retains uniform-reward degeneracy. Process supervision (c) combines human-labeled feedback \citep{lightman2023verify} or Math-Shepherd scores \citep{wang2024mathshepherd} with fixed weights. CALIPER (d) adapts these weights from estimated utility.}
\label{fig:intro}
\vspace{-12pt}
\end{figure}
Reference-assisted verification enriches answer grading but retains a scalar outcome reward, leaving stage-level credit assignment and uniform-reward degeneracy unresolved, as illustrated in Fig.~\ref{fig:intro}a,b. Process supervision \citep{uesato2022solving} provides finer feedback through human-annotated step labels \citep{lightman2023verify} or rollout-derived labels in Math-Shepherd \citep{wang2024mathshepherd}. However, a stage score quantifies alignment with a reference or grading criterion. It does not determine the supervision weight appropriate to that stage. Combining such scores with predefined weights leaves reward allocation insensitive to changes in the relationship between stage performance and answer correctness. Fig.~\ref{fig:intro}c illustrates this gap in adaptive credit allocation across perception, decomposition and inference despite access to intermediate feedback.

Stage utility guides allocation through the association between stage performance and answer correctness under the current rollout distribution. Low reference coverage does not establish the expected benefit of increasing a stage's reward weight. Allocation therefore requires an estimate of this association, conditioned on the policy's evolving behavior across spatial competencies. Rollout groups with variation in answer correctness supply the contrast needed for estimation, whereas uniformly correct or incorrect groups provide no such contrast. Adaptive supervision must estimate utility from informative groups and maintain reliable allocation as they become scarce.

To address this gap, we introduce SpatialTrace and CALIPER, as illustrated in Fig.~\ref{fig:intro}d. SpatialTrace represents heterogeneous spatial tasks through a common schema for perception, decomposition and inference while retaining available reasoning from incomplete traces. CALIPER uses this representation to allocate process supervision according to estimated stage utility. It estimates the association between stage coverage and answer correctness from answer-discriminative rollout groups, then applies the resulting allocation to all groups, including uniformly failing ones. A closed-loop harness monitors estimation reliability and validation performance to guide checkpoint rollback, rollout diversification and stopping. Our contributions are as follows.
\begin{itemize}[itemsep=1pt,parsep=0pt]
\item We formulate process-reward allocation as a policy-dependent estimation problem, distinguishing stage quality from its association with answer correctness. Utility estimation is restricted to informative groups, while process supervision applies to all groups.
\item We construct SpatialTrace, a reasoning-augmented spatial dataset spanning heterogeneous spatial tasks, in which approximately $84\%$ of items carry complete three-stage traces and the remainder retain available reasoning without imputed stage labels.
\item We propose CALIPER, a closed-loop framework that estimates stage utility from answer-discriminative rollout groups, applies the resulting allocation to all groups, including degenerate groups, and audits that estimate through a rule-based harness.
\item Across eight benchmarks and three backbones, CALIPER attains the highest mean under both protocols and improves on fixed allocation across RL optimizers. Ablations support adaptive allocation and harness control, with gains at larger model and data scales.
\end{itemize}

\section{Related Work}

\paragraph{Spatial Reasoning in Vision-Language Models.}
Spatial reasoning in vision-language models advances through spatial supervision, visual search and outcome optimization. SpatialVLM \citep{chen2024spatialvlm} and SpatialRGPT \citep{cheng2024spatialrgpt} derive metric and relational supervision from 3D reconstruction or depth. Benchmarks include 3DSRBench \citep{ma2024threedsrbench} for 3D relations, SpatialEval \citep{wang2024spatialeval} for synthetic reasoning, MME-RealWorld \citep{zhang2024mmerealworld} for real-world scenes and RoboSpatial \citep{song2025robospatial} for embodied settings. V* \citep{wu2024v}, ZoomEye \citep{shen2025zoomeye} and DeepEyes \citep{zheng2026deepeyes} improve evidence acquisition through region selection, cropping and zooming. RLVR extends to multimodal reasoning in Vision-R1 \citep{huang2026vision}, R1-Zero-style spatial training \citep{liao2025improved} and video spatial reasoning in SpaceR \citep{ouyang2025spacer}. These efforts improve supervision, evidence acquisition and outcome optimization while leaving intermediate-stage credit allocation unresolved.

\paragraph{Process Supervision in RLVR.}
Chain-of-thought prompting \citep{wei2022cot} makes intermediate reasoning explicit, while DeepSeek-R1 \citep{guo2025deepseekr1} optimizes reasoning trajectories with verifiable outcome rewards. GRPO \citep{shao2024deepseekmath} uses group-relative advantages, DAPO \citep{yu2025dapo} modifies clipping and rollout filtering, and GSPO \citep{zheng2025gspo} uses sequence-level importance ratios. These methods govern how rewards are translated into policy updates, but do not change their granularity. Process supervision instead assigns feedback to intermediate reasoning steps, using direct annotations \citep{lightman2023verify} or labels derived from rollouts \citep{wang2024mathshepherd}. PRIME \citep{cui2025prime} integrates implicit process rewards into online RL, and PURE \citep{cheng2025pure} replaces summation with min-form credit assignment. Such feedback alleviates outcome-reward sparsity, but its aggregation follows a fixed rule, leaving reward allocation insensitive to the policy's changing bottlenecks. CALIPER addresses this orthogonal problem by estimating policy-dependent stage utility and adapting process-reward weights accordingly.

\section{SpatialTrace Dataset}
\label{sec:data}

\paragraph{Task Formulation.}
SpatialTrace targets spatial question answering with an explicit reasoning trace. Given an image $I$ and a question $x$, a vision-language policy receives $q=(I,x)$ and generates a trace $\tau$ followed by an answer $a$. The desired behavior comprises extracting relevant visual evidence, decomposing the question into spatial subproblems, and deriving an answer consistent with that evidence. We represent these components as $\tau=(\tau_{\mathrm p},\tau_{\mathrm d},\tau_{\mathrm i})$, corresponding to perception, decomposition and inference. A reference answer $a^\star$ supports outcome verification, while stage annotations support separate assessment of these reasoning components. This distinction permits analysis of intermediate reasoning even when the final answer is incorrect.
\begin{figure}[tbh]
\centering
\vspace{-8pt}
\includegraphics[width=\linewidth]{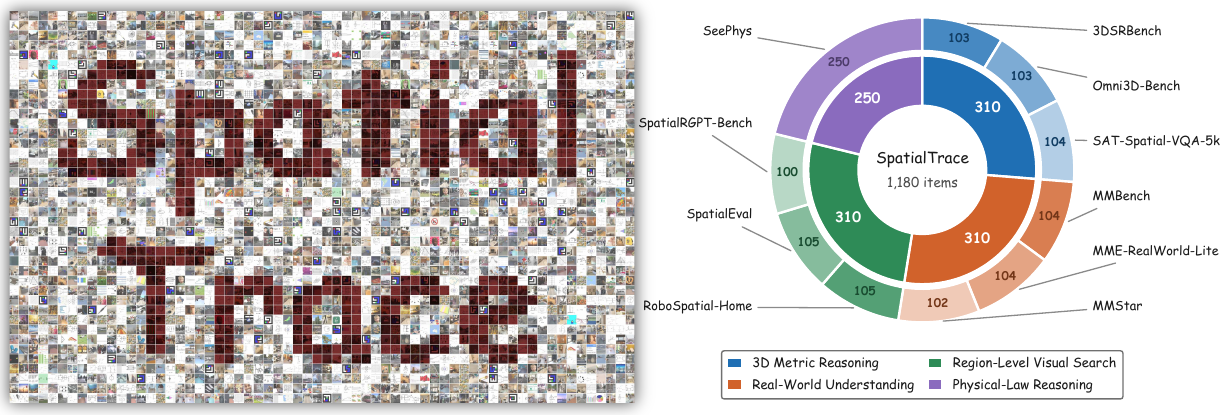}%
\caption{SpatialTrace examples (left) and source composition (right).}
\label{fig:corpus}
\vspace{-8pt}
\end{figure}

\paragraph{Construction.}
Learning this behavior requires coverage of spatial problems with different perceptual and inferential demands. SpatialTrace draws metric 3D questions from 3DSRBench \citep{ma2024threedsrbench}, Omni3D-Bench \citep{brazil2023omni3d} and SAT-Spatial-VQA-5k \citep{ray2025satdynamicspatialaptitude}. Open-domain questions are sampled from MMBench \citep{liu2023mmbench}, MME-RealWorld-Lite \citep{zhang2024mmerealworld} and MMStar \citep{chen2024mmstar}. Region-level search instances are sampled from RoboSpatial-Home \citep{song2025robospatial}, SpatialEval \citep{wang2024spatialeval} and SpatialRGPT-Bench \citep{cheng2024spatialrgpt}, with SeePhys \citep{xiang2025seephys} supplying auxiliary physical reasoning. Fig.~\ref{fig:corpus} summarizes SpatialTrace's $1{,}180$ single-image questions with verifiable answers. Family quotas preserve coverage, with source-stratified splits of $1{,}000$ training and $180$ validation items.

\paragraph{Traces.}
The annotations encode the relationship between visual evidence, intermediate reasoning and the reference answer. Perception records relevant visual observations. Decomposition identifies the subproblems required by the question. Inference connects the observations to the answer. Each record also includes answer reconciliation and the final answer, allowing discrepancies between visual evidence and the source label to be recorded. An image-grounded annotator drafts the trace, and a text-only editor refines its reasoning and records unsupported inferences or ambiguity. Approximately $84\%$ of the corpus contains all three stage fields. Other records retain their answers and available reasoning without imputed stage labels. Appendix~\ref{app:data} details the annotation protocol.

\section{Methodology}
\label{sec:method}

\paragraph{Problem Formulation.}
Given an image--question pair $q=(I,x)$ and a verifiable answer $a^\star$, a policy $\pi_\theta$ generates a group $g$ of $n$ rollouts $\{o_i\}_{i\in g}\sim\pi_\theta(\cdot\mid q)$. Let $r_i$ denote the reward of rollout $o_i$. Group-relative policy optimization \citep{shao2024deepseekmath} computes its advantage from the group mean $\mu_g$ and standard deviation $\sigma_g$,
\begin{equation}
A_i=\begin{cases}(r_i-\mu_g)/\sigma_g,&\sigma_g>0,\\0,&\sigma_g=0,\end{cases}\qquad
\mu_g=\tfrac1n\textstyle\sum_{j\in g} r_j,\;\; \sigma_g^2=\tfrac1n\textstyle\sum_{j\in g} (r_j-\mu_g)^2 ,
\label{eq:grpo}
\end{equation}
The same $A_i$ applies to every token of $o_i$. If all rewards in $g$ agree, then $\sigma_g=0$ and the advantages are set to zero. Such groups contribute no advantage-weighted policy gradient, although other objective terms may remain. Let $\mathrm{acc}_i\in\{0,1\}$ denote the verifier judgment of the answer in $o_i$ against $a^\star$. Under a binary outcome reward, groups that are uniformly correct or uniformly incorrect both yield zero advantages. The problem is to assign credit among perception, decomposition and inference even when answer-level rewards cannot distinguish the rollouts.

\begin{wrapfigure}{r}{0.58\textwidth}
\centering
\vspace{-16pt}
\includegraphics[width=\linewidth]{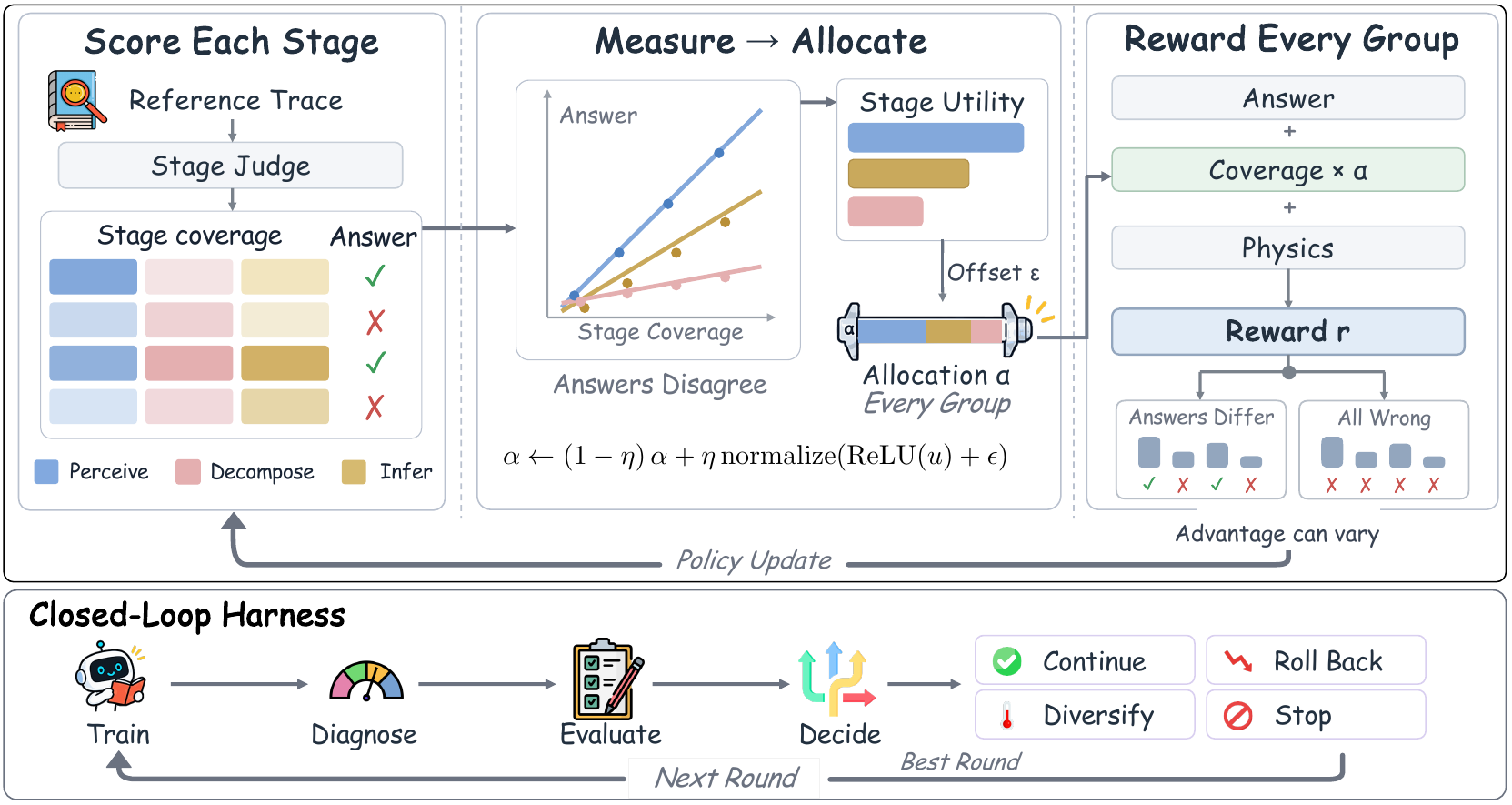}%
\caption{CALIPER scores stages against reference traces, estimates utility from groups with mixed answer correctness, and updates the policy using rewards for every group. Validation and estimator diagnostics guide continuation, rollback, rollout diversification or stopping.}
\label{fig:method}
\vspace{-18pt}
\end{wrapfigure}
\paragraph{Overview.}
Fig.~\ref{fig:method} shows CALIPER's two coupled loops. During a policy update, a stage judge scores each rollout against the corresponding SpatialTrace reference for perception, decomposition and inference. Groups with both correct and incorrect answers support the stage-utility estimate. The estimated utilities determine the weights used to reward every group. Across training rounds, the harness checks validation performance and estimator reliability. It then continues training, restores a checkpoint, increases rollout diversity or stops.

\paragraph{Stage-Decomposed Process Reward.}
Let $\mathcal S$ index the perception, decomposition and inference stages. The reference trace for $q$ is $\tau^\star=(\tau^\star_s)_{s\in\mathcal S}$. A judge $J$ assigns each rollout a reference-coverage score $\mathrm{cov}_s(o_i)=J(o_i,\tau^\star_s)\in[0,1]$ for stage $s$. Let $\mathrm{phys}_i\in[0,1]$ be a physical-plausibility score and $\mathrm{len}_i\in[0,1]$ a response-length penalty. The nonnegative weights $w_a$, $w_m$, $w_p$ and $w_\ell\in[0,1]$ control answer correctness, process coverage, physical plausibility and length, respectively. The allocation $\alpha\in\Delta^{|\mathcal S|}$ lies on the probability simplex and distributes the process weight $w_m$ across stages. We define the rollout reward as
\begin{equation}
r(o_i)\;=\;\Bigl[\,w_a\,\mathrm{acc}_i \;+\; w_m\!\!\sum_{s\in\mathcal S}\!\alpha_s\,\mathrm{cov}_s(o_i)
\;+\; w_p\,\mathrm{phys}_i\,\Bigr]\cdot\bigl(1-w_\ell\,\mathrm{len}_i\bigr),
\label{eq:reward}
\end{equation}
The score $\mathrm{phys}_i$ quantifies consistency with physical constraints on magnitudes, units and relations. For response length $|o_i|$ in tokens, target length $L_0$ and cap $L_{\max}>L_0$, the penalty is
\[
\mathrm{len}_i=\min\!\left\{1,\max\!\left\{0,\frac{|o_i|-L_0}{L_{\max}-L_0}\right\}\right\}.
\]
It increases linearly from zero at $L_0$ to one at $L_{\max}$. For incomplete references, the reward sums over available stages and renormalizes their weights.

\paragraph{Stage Utility.}
Stage utility is the conditional association between coverage and answer correctness under the current policy. Among groups with complete references, only answer-discriminative groups $\mathcal M=\{g:0<\sum_{i\in g}\mathrm{acc}_i<n\}$ contain the outcome variation needed to estimate it. For rollout $o_i$ in such a group, we center each stage score and the outcome within the group,
\[
\tilde{x}_{i,s}=\mathrm{cov}_s(o_i)-\frac1n\sum_{j\in g}\mathrm{cov}_s(o_j),
\qquad
\tilde{y}_i=\mathrm{acc}_i-\frac1n\sum_{j\in g}\mathrm{acc}_j.
\]
The vector $\tilde{x}_i$ collects the centered stage scores. The design matrix $\tilde X$ stacks these vectors as rows, and $\tilde y$ stacks the corresponding outcomes. With $\bar\sigma^2=\mathrm{tr}(\tilde X^{\!\top}\tilde X)/|\mathcal S|$ and ridge coefficient $\lambda>0$, we estimate utility $u\in\mathbb R^{|\mathcal S|}$ by
\begin{equation}
u=\arg\min_{\beta\in\mathbb R^{|\mathcal S|}}\left\{
\sum_{g\in\mathcal M}\sum_{i\in g}
\bigl(\tilde y_i-\beta^{\!\top}\tilde x_i\bigr)^2
+\lambda\bar\sigma^2\lVert\beta\rVert_2^2\right\}
=\bigl(\tilde X^{\!\top}\tilde X+\lambda\bar\sigma^2\mathrm I\bigr)^{-1}
\tilde X^{\!\top}\tilde y,
\label{eq:utility}
\end{equation}
where $\mathrm I$ is the identity matrix and the closed form applies when $\bar\sigma^2>0$. Centering compares rollouts of the same question, and joint regression accounts for correlations among stage scores. The coefficients quantify each stage's association with correctness after accounting for the other stages. The scale $\bar\sigma^2$ is the mean diagonal of the Gram matrix, so the penalty follows the feature scale. Across batches, exponential decay $\gamma\in(0,1)$ gives recent samples greater weight in the sufficient statistics $\tilde X^{\!\top}\tilde X$ and $\tilde X^{\!\top}\tilde y$. Let $N_{\mathrm{eff}}$ be the corresponding decayed sample mass and $N_{\min}$ its update threshold. We retain the current allocation unless $N_{\mathrm{eff}}\ge N_{\min}$, feature variation is nonzero and at least one coefficient is positive.

\paragraph{Utility-Adaptive Allocation.}\label{sec:method-alloc}
Define $\operatorname{ReLU}(u)=\max(u,0)$ elementwise and $\mathrm{normalize}(v)=v/\sum_s v_s$. An offset $\epsilon>0$ keeps the normalized target positive, and a step size $\eta\in(0,1]$ limits the change between updates. CALIPER interpolates between the current allocation and this target,
\begin{equation}
\alpha \;\leftarrow\; (1-\eta)\,\alpha \;+\; \eta\,\mathrm{normalize}\bigl(\operatorname{ReLU}(u)+\epsilon\bigr),
\label{eq:alloc}
\end{equation}
The offset is added componentwise. It preserves positive weights but does not impose a lower bound of $\epsilon$ after normalization. Utilities are estimated from answer-discriminative groups, and the resulting allocation supervises all groups.

\paragraph{Closed-Loop Harness.}
As training progresses, answer-discriminative groups may become scarce, weakening the utility estimate even when the allocation appears stable. Validation performance can also deteriorate during training. The harness diagnoses both failure modes to determine whether and how training should continue. At round $k$, let $\alpha_k$ be the current allocation, $m_k$ the validation accuracy, $\sigma$ its standard error and $m^\star$ the best accuracy from earlier rounds. We record allocation drift $\delta_k=\lVert\alpha_k-\alpha_{k-1}\rVert_1$ and the mean number $n^{\mathcal M}_k$ of answer-discriminative groups per batch. The allocation target is $\bar u_k=\mathrm{normalize}(\operatorname{ReLU}(u_k)+\epsilon)$, with separation $\Delta u_k=\max_s\bar u_{k,s}-\min_s\bar u_{k,s}$. The decision $\mathcal D_k$ applies the fixed regression tolerance $\rho_{\mathrm r}$, drift threshold $\epsilon_\delta$, group-count threshold $\nu_{\min}$ and separation threshold $\epsilon_u$ in the following order,
\begin{equation}
\mathcal{D}_k=\begin{cases}
\textsc{rollback} & m_k<m^\star-\max(\rho_{\mathrm r}\,m^\star,\,2\sigma),\\[2pt]
\textsc{diversify} & \delta_k<\epsilon_\delta\ \wedge\ n^{\mathcal M}_k<\nu_{\min},\\[2pt]
\textsc{stop} & \delta_k<\epsilon_\delta\ \wedge\ n^{\mathcal M}_k\ge\nu_{\min}\ \wedge\ \max(\Delta u_k,\Delta u_{k-1})<\epsilon_u,\\[2pt]
\textsc{continue} & \text{otherwise},
\end{cases}
\label{eq:harness}
\end{equation}
Rollback requires a saved checkpoint. We set $\Delta u_0=+\infty$ and use $\Delta u_k=+\infty$ when the utility estimate is unavailable, disabling utility-based stopping in those rounds. \textsc{rollback} restores the policy, allocation and rollout temperature of the best validation round. \textsc{diversify} raises temperature when allocation drift is low and answer-discriminative groups are scarce. \textsc{stop} terminates training when current drift is low, sufficient answer-discriminative groups are available and utility separation is small in two consecutive rounds. Evaluation uses the best validation checkpoint.

\paragraph{Policy Objective.}
The policy update computes $A_i$ from the stage-weighted reward in Eq.~\ref{eq:reward}. Let $o_{i,t}$ be token $t$ of rollout $o_i$, $o_{i,<t}$ its prefix and $\theta_{\mathrm{old}}$ the parameters of the sampling policy. The token-level importance ratio is
\[
\rho_{i,t}(\theta)=
\frac{\pi_\theta(o_{i,t}\mid q,o_{i,<t})}
{\pi_{\theta_{\mathrm{old}}}(o_{i,t}\mid q,o_{i,<t})}.
\]
With lower and upper clipping tolerances $\varepsilon_{\mathrm{lo}},\varepsilon_{\mathrm{hi}}>0$, the clipped policy-loss term averaged over sampled questions and rollout groups is
\begin{equation}
\mathcal{L}_{\mathrm{policy}}(\theta)\;=\;-\,\mathbb{E}\Bigl[\tfrac{1}{n}\textstyle\sum_i \tfrac{1}{|o_i|}\sum_t
\min\bigl(\rho_{i,t}A_i,\;\mathrm{clip}(\rho_{i,t},1-\varepsilon_{\text{lo}},1+\varepsilon_{\text{hi}})A_i\bigr)\Bigr],
\label{eq:obj}
\end{equation}
For GRPO, we augment this objective with a KL penalty that constrains deviation from the frozen reference policy. The complete CALIPER training procedure is given in Algorithm~\ref{alg:caliper}.

\begin{proposition}[Stage-Contrast Preservation]
\label{prop:stage-contrast}
Fix the policy and the conditional distributions of complete-reference groups. Let $d=|\mathcal S|$, $p\in(0,1)$ be the rollout fraction in answer-discriminative groups, $D$ the centered feature covariance in the remaining groups, and $c$ the centered feature--outcome cross-moment in answer-discriminative groups. Assume exact moments, positive feature variance in both group types, and $\lambda,\epsilon>0$. Let $\mathcal T_\epsilon$ denote the target map in Eq.~\ref{eq:alloc} and $\alpha_{\mathrm{unif}}$ the uniform allocation. The mixed-group utility $u_{\mathrm M}$ and its target are independent of $p$, whereas applying the same estimator to all groups gives a utility $u_p$ satisfying
\begin{equation}
\bigl\|\mathcal T_\epsilon(u_p)-\alpha_{\mathrm{unif}}\bigr\|_1
\le \frac{2\sqrt d\,p\|c\|_2}{(1-p)\lambda\epsilon\operatorname{tr}D}
\longrightarrow 0 \quad\text{as }p\downarrow0.
\label{eq:contrast-bound}
\end{equation}
Moreover, a bounded three-stage construction preserves a positive association between normalized advantages and a correctness-associated stage contrast under mixed-group estimation, while this association vanishes under pooled estimation as $p\downarrow0$. The proof is provided in Appendix~\ref{app:theory}.
\end{proposition}

\section{Experiments}
\label{sec:exp}

\begin{table}[tbh]
\centering
\renewcommand{\arraystretch}{1.15}
\setlength{\tabcolsep}{3.5pt}
\caption{Main results (\%). \textbf{Avg} is the mean, and \textbf{1st} counts wins (ties included). \textbf{Bold}/\underline{underline} mark best/second-best scores. $\Delta$ compares protocols.}
\label{tab:main}
\resizebox{\textwidth}{!}{%
\begin{tabular}{lcccccccccccc}
\toprule
\textbf{Method} & \textbf{Harness} & \textsc{3DSR} & \textsc{MMB} & \textsc{MME-RW} & \textsc{MMStar} & \textsc{Omni3D} & \textsc{RoboSp} & \textsc{SAT} & \textsc{SpatEval} & \textbf{Avg} & $\bm{\Delta}$ & \textbf{1st} \\
\midrule
\multicolumn{13}{l}{{\bfseries Qwen3-VL-4B}~\normalfont \citep{bai2025qwen3vltechnicalreport}} \\[1pt]
Direct (No RL) & \textcolor{black!35}{\ding{55}} & 55.00 & 89.00 & 49.00 & 61.00 & \bfseries 38.90 & 43.92 & 75.00 & 56.00 & 58.48 & \textcolor{black!35}{\ding{55}} & 1 \\
CoT (No RL) & \textcolor{black!35}{\ding{55}} & 59.00 & 92.00 & 47.50 & 67.00 & 31.85 & 44.72 & 73.50 & 68.50 & 60.51 & \textcolor{black!35}{\ding{55}} & 0 \\
\hdashline[0.5pt/1.5pt]
 & \textcolor{black!35}{\ding{55}} & 60.50 & 91.00 & 46.00 & 66.50 & 33.70 & 46.15 & 76.00 & 70.00 & 61.23 &  & 0 \\
\multirow{-2}{*}{Outcome Reward} & \textcolor{okgreen}{\ding{51}} & 58.00 & 91.00 & 50.50 & 64.50 & 33.65 & 46.72 & 76.00 & 70.50 & 61.36 & \multirow{-2}{*}{$+0.13$} & 0 \\
\hdashline[0.5pt/1.5pt]
 & \textcolor{black!35}{\ding{55}} & 61.00 & 92.00 & 47.50 & 68.00 & 33.95 & 47.13 & \bfseries 77.00 & 68.00 & 61.82 &  & 1 \\
\multirow{-2}{*}{Reference-Assisted} & \textcolor{okgreen}{\ding{51}} & 61.50 & 92.00 & 50.00 & 65.00 & 32.55 & 48.03 & 75.00 & 71.50 & 61.95 & \multirow{-2}{*}{$+0.13$} & 0 \\
\hdashline[0.5pt/1.5pt]
 & \textcolor{black!35}{\ding{55}} & 63.00 & \bfseries 93.50 & 49.50 & 68.50 & \underline{37.20} & \underline{49.20} & 74.00 & 69.50 & 63.05 &  & 1 \\
\multirow{-2}{*}{Process Reward Model} & \textcolor{okgreen}{\ding{51}} & 63.50 & 92.50 & 48.00 & \underline{69.50} & 37.05 & 46.11 & 75.00 & 71.50 & 62.90 & \multirow{-2}{*}{\textcolor{black}{$-0.15$}} & 0 \\
\hdashline[0.5pt/1.5pt]
\rowcolor{ourrow} & \textcolor{black!35}{\ding{55}} & \bfseries 66.00 & \bfseries 93.50 & \underline{52.00} & 69.00 & 37.05 & 46.95 & \underline{76.50} & \underline{72.50} & \underline{64.19} &  & 2 \\
\rowcolor{ourrow}\multirow{-2}{*}{CALIPER~(ours)} & \textcolor{okgreen}{\ding{51}} & \underline{65.50} & \underline{93.00} & \bfseries 52.50 & \bfseries 70.00 & 36.35 & \bfseries 50.37 & 72.50 & \bfseries 75.00 & \bfseries 64.40 & +0.21 & \bfseries 4 \\
\addlinespace[2pt]\midrule
\multicolumn{13}{l}{{\bfseries Qwen2.5-VL-7B}~\normalfont \citep{bai2025qwen25vltechnicalreport}} \\[1pt]
Direct (No RL) & \textcolor{black!35}{\ding{55}} & 49.50 & 88.00 & 44.50 & 62.00 & 23.15 & 43.73 & 51.50 & 53.00 & 51.92 & \textcolor{black!35}{\ding{55}} & 0 \\
CoT (No RL) & \textcolor{black!35}{\ding{55}} & 51.50 & 88.50 & 45.50 & \underline{65.00} & \underline{24.65} & \underline{47.23} & 54.50 & 57.00 & 54.23 & \textcolor{black!35}{\ding{55}} & 0 \\
\hdashline[0.5pt/1.5pt]
 & \textcolor{black!35}{\ding{55}} & \underline{55.00} & 89.00 & \underline{51.00} & 62.50 & 22.75 & 44.05 & 58.50 & 54.50 & 54.66 &  & 0 \\
\multirow{-2}{*}{Outcome Reward} & \textcolor{okgreen}{\ding{51}} & \bfseries 55.50 & \underline{89.50} & 49.50 & \bfseries 65.50 & 24.05 & 43.06 & \bfseries 62.50 & 55.50 & 55.64 & \multirow{-2}{*}{$+0.98$} & 3 \\
\hdashline[0.5pt/1.5pt]
 & \textcolor{black!35}{\ding{55}} & 53.50 & \underline{89.50} & \bfseries 52.00 & \bfseries 65.50 & 20.45 & 44.10 & 57.50 & 56.50 & 54.88 &  & 2 \\
\multirow{-2}{*}{Reference-Assisted} & \textcolor{okgreen}{\ding{51}} & 53.00 & 88.50 & 50.00 & 63.50 & 22.45 & 43.47 & \underline{62.00} & 57.00 & 54.99 & \multirow{-2}{*}{$+0.11$} & 0 \\
\hdashline[0.5pt/1.5pt]
 & \textcolor{black!35}{\ding{55}} & 52.50 & 85.50 & 47.00 & 59.50 & \bfseries 24.75 & 45.25 & 53.50 & 54.50 & 52.81 &  & 1 \\
\multirow{-2}{*}{Process Reward Model} & \textcolor{okgreen}{\ding{51}} & 48.50 & 85.50 & 44.50 & 60.00 & 24.00 & 45.87 & 56.00 & 55.50 & 52.48 & \multirow{-2}{*}{\textcolor{black}{$-0.33$}} & 0 \\
\hdashline[0.5pt/1.5pt]
\rowcolor{ourrow} & \textcolor{black!35}{\ding{55}} & 54.50 & \bfseries 90.50 & 50.50 & 63.50 & 23.05 & 46.73 & 57.00 & \underline{62.00} & \underline{55.97} &  & 1 \\
\rowcolor{ourrow}\multirow{-2}{*}{CALIPER~(ours)} & \textcolor{okgreen}{\ding{51}} & 53.50 & \bfseries 90.50 & 49.50 & \bfseries 65.50 & \bfseries 24.75 & \bfseries 48.27 & 55.50 & \bfseries 62.50 & \bfseries 56.25 & +0.28 & \bfseries 5 \\
\addlinespace[2pt]\midrule
\multicolumn{13}{l}{{\bfseries InternVL3-8B}~\normalfont \citep{zhu2025internvl3}} \\[1pt]
Direct (No RL) & \textcolor{black!35}{\ding{55}} & 51.00 & \bfseries 89.00 & 54.00 & 71.00 & 36.95 & 39.90 & \underline{74.50} & 50.00 & 58.29 & \textcolor{black!35}{\ding{55}} & 1 \\
CoT (No RL) & \textcolor{black!35}{\ding{55}} & 56.50 & \underline{88.50} & 53.00 & \bfseries 73.00 & 38.65 & 44.32 & 72.50 & 50.50 & 59.62 & \textcolor{black!35}{\ding{55}} & 1 \\
\hdashline[0.5pt/1.5pt]
 & \textcolor{black!35}{\ding{55}} & 55.50 & 88.00 & 53.00 & 67.50 & 37.40 & 41.63 & \bfseries 75.00 & 52.50 & 58.82 &  & 1 \\
\multirow{-2}{*}{Outcome Reward} & \textcolor{okgreen}{\ding{51}} & 55.50 & 87.50 & 54.50 & 66.50 & 37.00 & 43.75 & 74.00 & \underline{57.50} & 59.53 & \multirow{-2}{*}{$+0.71$} & 0 \\
\hdashline[0.5pt/1.5pt]
 & \textcolor{black!35}{\ding{55}} & 56.50 & 88.00 & 52.50 & 71.50 & 37.90 & 44.28 & 74.00 & 54.00 & 59.84 &  & 0 \\
\multirow{-2}{*}{Reference-Assisted} & \textcolor{okgreen}{\ding{51}} & 55.00 & 88.00 & 54.00 & 68.50 & 38.10 & 47.19 & 74.00 & \underline{57.50} & 60.29 & \multirow{-2}{*}{$+0.45$} & 0 \\
\hdashline[0.5pt/1.5pt]
 & \textcolor{black!35}{\ding{55}} & \underline{57.50} & \underline{88.50} & \bfseries 56.50 & 70.00 & 38.15 & 49.08 & 73.50 & 56.00 & 61.15 &  & 1 \\
\multirow{-2}{*}{Process Reward Model} & \textcolor{okgreen}{\ding{51}} & 57.00 & \bfseries 89.00 & 52.50 & 70.00 & 38.35 & \underline{50.04} & 72.50 & 54.50 & 60.49 & \multirow{-2}{*}{\textcolor{black}{$-0.67$}} & 1 \\
\hdashline[0.5pt/1.5pt]
\rowcolor{ourrow} & \textcolor{black!35}{\ding{55}} & \bfseries 58.00 & \bfseries 89.00 & 55.00 & \underline{72.50} & \bfseries 41.65 & 46.10 & \bfseries 75.00 & \underline{57.50} & \underline{61.84} &  & \bfseries 4 \\
\rowcolor{ourrow}\multirow{-2}{*}{CALIPER~(ours)} & \textcolor{okgreen}{\ding{51}} & \underline{57.50} & \bfseries 89.00 & \underline{55.50} & 71.00 & \underline{40.25} & \bfseries 52.14 & \bfseries 75.00 & \bfseries 60.00 & \bfseries 62.55 & +0.71 & \bfseries 4 \\
\bottomrule
\end{tabular}}
\vspace{-12pt}
\end{table}

\paragraph{Implementation Details.}
The primary experiments use the GRPO implementation of EasyR1 \citep{zheng2025easyr1} on verl \citep{sheng2025hybridflow}. The backbone is frozen throughout and a LoRA adapter \citep{hu2022lora} of rank $64$ is trained with $n=8$ rollouts per question at temperature $1.2$, AdamW at learning rate $1{\times}10^{-5}$, and $512$ questions per optimizer step. Backbones are Qwen3-VL-4B \citep{bai2025qwen3vltechnicalreport}, Qwen2.5-VL-7B \citep{bai2025qwen25vltechnicalreport} and InternVL3-8B \citep{zhu2025internvl3}. One-shot training comprises ten optimizer steps, approximately five passes over $1{,}000$ items. Harness training comprises at most six rounds of five steps. Each run uses eight NVIDIA H200 GPUs (141 GB each) on a single node. Unless varied explicitly, the backbones share the reward, estimator, allocation and harness settings specified in Appendix~\ref{app:impl}.
\paragraph{Baselines.}
We compare two untrained configurations and three reward-based training baselines. Direct prompting requests an answer without a reasoning trace. Chain-of-thought prompting \citep{wei2022cot} uses the same three-stage format as trained methods to isolate explicit reasoning before training. Within each training protocol, all methods share the backbone, adapter and rollout settings. Outcome reward uses final-answer supervision, following GRPO \citep{shao2024deepseekmath} and Vision-R1 \citep{huang2026vision}. Reference-assisted reward is a controlled variant in which the verifier receives the SpatialTrace reference as grading context. The process reward model (PRM) baseline follows stage-level labeling in \citet{lightman2023verify} and \citet{wang2024mathshepherd}, and combines its three predicted scores under fixed weights. Appendix~\ref{app:prompts} details reward scorers, baseline and prompts.

\paragraph{Benchmarks \& Metrics.}
We train each backbone on our SpatialTrace dataset (Section~\ref{sec:data}) and evaluate it on eight primary benchmarks. Four target 3D and relational spatial understanding, namely 3DSRBench \citep{ma2024threedsrbench}, RoboSpatial-Home \citep{song2025robospatial}, and spatial question-answering splits built on Omni3D \citep{brazil2023omni3d} and SAT \citep{ray2025satdynamicspatialaptitude}. Two are the general multimodal benchmarks MMBench \citep{liu2023mmbench} and MMStar \citep{chen2024mmstar}, and the remaining two are the high-resolution real-world benchmark MME-RealWorld-Lite \citep{zhang2024mmerealworld} and the synthetic spatial puzzles of SpatialEval \citep{wang2024spatialeval}. Training and evaluation images are disjoint (Appendix~\ref{app:data}). We report benchmark scores (\%) and their unweighted mean (Avg) on fixed $200$-item subsets. The metrics include binary accuracy, a mixed numerical score for Omni3D, and a proxy context score within RoboSpatial. Appendix~\ref{app:impl} specifies these differences.

\subsection{Main Results}
\paragraph{Performance Comparison.}
Table~\ref{tab:main} shows that CALIPER achieves the highest mean score across the eight benchmarks for all three backbones under both training protocols. Under harness training, its margin over the strongest trained baseline reaches $2.06$ percentage points on InternVL3-8B. PRM improves on outcome reward for two backbones but falls behind on Qwen2.5-VL-7B, indicating that intermediate supervision alone is insufficient. The consistent advantage over frozen allocation supports adapting stage weights to the current policy.

\begin{wraptable}{R}{0.62\textwidth}
\centering
\vspace{-18pt}
\caption{Qwen3-VL-4B OOD accuracy (\%).}
\label{tab:ood}
\resizebox{.62\columnwidth}{!}{
\renewcommand{\arraystretch}{1.18}
\begin{tabular}{@{}lccccc@{}}
\toprule
\textbf{Method} & BLINK & \shortstack{RealWorld\\QA} & \shortstack{Emb\\Spatial} & \shortstack{View\\Spatial} & Avg \\
\midrule
Direct (no RL) & 80.50 & 73.00 & 76.50 & 37.50 & 66.88 \\
CoT (no RL) & 83.00 & 75.00 & 69.50 & 43.00 & 67.63 \\
Outcome Reward & 84.00 & 71.50 & 72.50 & 41.50 & 67.38 \\
Reference-Assisted & 85.00 & 71.50 & 73.00 & 43.00 & 68.13 \\
Process Reward Model & 80.50 & 72.00 & 72.00 & 45.00 & 67.38 \\
Frozen $\alpha$ & 83.00 & 74.50 & 73.00 & 41.50 & 68.00 \\
\rowcolor{ourrow}CALIPER (ours) & 85.50 & 76.00 & 76.50 & 46.00 & 71.00 \\
\midrule
Harness PRM & 81.50 & 76.00 & 75.00 & 45.50 & 69.50 \\
\bottomrule
\end{tabular}}
\vspace{-2pt}
\end{wraptable}

\paragraph{OOD Generalization.}
Table~\ref{tab:ood} shows CALIPER's strongest relative gain on viewpoint-dependent spatial reasoning. Its mean exceeds frozen allocation by $3.00$ percentage points, with the largest individual gain on ViewSpatial-Bench. The evaluation covers BLINK-Spatial \citep{fu2024blink}, RealWorldQA \citep{xai2024realworldqa}, EmbSpatial-Bench \citep{u2024embspatial} and ViewSpatial-Bench \citep{li2025viewspatial} under fixed prompts and decoding after SpatialTrace training. Harness PRM uses the separate harness protocol. The pattern is consistent with better transfer of stage-specific supervision to viewpoint reasoning while retaining performance on the other OOD tasks.

\subsection{Framework Analysis}

\begin{figure}[tbh]
\centering
\begin{minipage}[t]{0.53\linewidth}
\vspace{0pt}
\centering
\captionof{table}{Optimizer transfer under harness training, with $\Delta$ relative to CALIPER.}
\label{tab:algo}
\vspace{3pt}
\resizebox{\linewidth}{!}{%
\begin{tabular}{llcccccc}
\toprule
\textbf{Optim.} & \textbf{Method} & \multicolumn{2}{c}{\textsc{Q3-VL-4B}} & \multicolumn{2}{c}{\textsc{Q2.5-VL-7B}} & \multicolumn{2}{c}{\textsc{IVL3-8B}} \\
\cmidrule(lr){3-4}\cmidrule(lr){5-6}\cmidrule(lr){7-8}
 & & Avg & $\Delta$ & Avg & $\Delta$ & Avg & $\Delta$ \\
\midrule
GRPO & Outcome & 61.36 & -3.04& 55.64 & -0.61& 59.53 & -3.02 \\
 & Ref.-assisted & 61.95 & -2.45& 54.99 & -1.26& 60.29 & -2.26 \\
 & Frozen $\alpha$ & 62.47 & -1.93& 55.22 & -1.03& 60.96 & -1.59 \\
\rowcolor{ourrow} & CALIPER & 64.40 & \textemdash & 56.25 & \textemdash & 62.55 & \textemdash \\
\midrule
DAPO & Outcome & 63.31 & $-0.04$ & 56.31 & $+2.34$ & 61.34 & $-1.13$ \\
 & Ref.-assisted & 63.36 & $+0.01$ & 56.26 & $+2.29$ & 61.21 & $-1.27$ \\
 & Frozen $\alpha$ & 62.56 & $-0.79$ & 53.38 & $-0.59$ & 61.09 & $-1.39$ \\
\rowcolor{ourrow} & CALIPER & 63.35 & \textemdash & 53.97 & \textemdash & 62.48 & \textemdash \\
\midrule
GSPO & Outcome & 61.43 & $-1.71$ & 55.46 & $+0.78$ & 59.97 & $-0.61$ \\
 & Ref.-assisted & 62.95 & $-0.20$ & 53.91 & $-0.76$ & 60.17 & $-0.42$ \\
 & Frozen $\alpha$ & 62.78 & $-0.37$ & 52.75 & $-1.92$ & 60.44 & $-0.14$ \\
\rowcolor{ourrow} & CALIPER & 63.15 & \textemdash & 54.67 & \textemdash & 60.58 & \textemdash \\
\bottomrule
\end{tabular}%
}
\end{minipage}%
\hfill
\begin{minipage}[t]{0.45\linewidth}
\vspace{10.25pt}
\centering
\includegraphics[width=\linewidth]{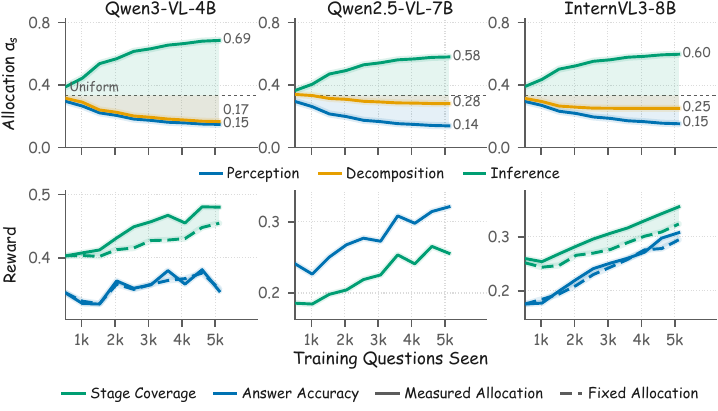}\par
\vspace{-7.5pt}
\captionof{figure}{Stage allocation (top) and rewards (bottom) over optimizer steps. Dashed curves use fixed allocation.}
\label{fig:alpha}
\end{minipage}
\vspace{-4pt}
\end{figure}

\paragraph{Generality Across Optimizers.}
Table~\ref{tab:algo} shows that adaptive allocation improves performance across distinct policy optimizers. Under both DAPO \citep{yu2025dapo} and GSPO \citep{zheng2025gspo}, CALIPER outperforms frozen allocation, with gains of up to $1.92$ percentage points despite their different update rules. On Qwen2.5-VL-7B, outcome reward remains stronger than process-based variants, indicating that adaptive allocation improves process supervision without requiring process rewards to universally dominate outcome rewards. These comparisons support utility-based allocation across the tested policy objectives.

\paragraph{Reward Allocation Dynamics.}
Fig.~\ref{fig:alpha} shows a consistent reallocation toward inference across backbones. Starting from uniform weights, inference accounts for $0.58$--$0.69$ of the final process-reward allocation. The lower panels further reveal that higher stage coverage does not uniformly translate into higher answer accuracy: coverage increases on Qwen3-VL-4B despite fluctuating accuracy, whereas the other two backbones exhibit more aligned improvements in both. The fixed-allocation baselines also follow backbone-specific trajectories on Qwen3-VL-4B and InternVL3-8B. These trajectories support using within-question correctness contrasts to estimate stage utility.

\begin{figure}[tbh]
\centering
\includegraphics[width=.985\textwidth]{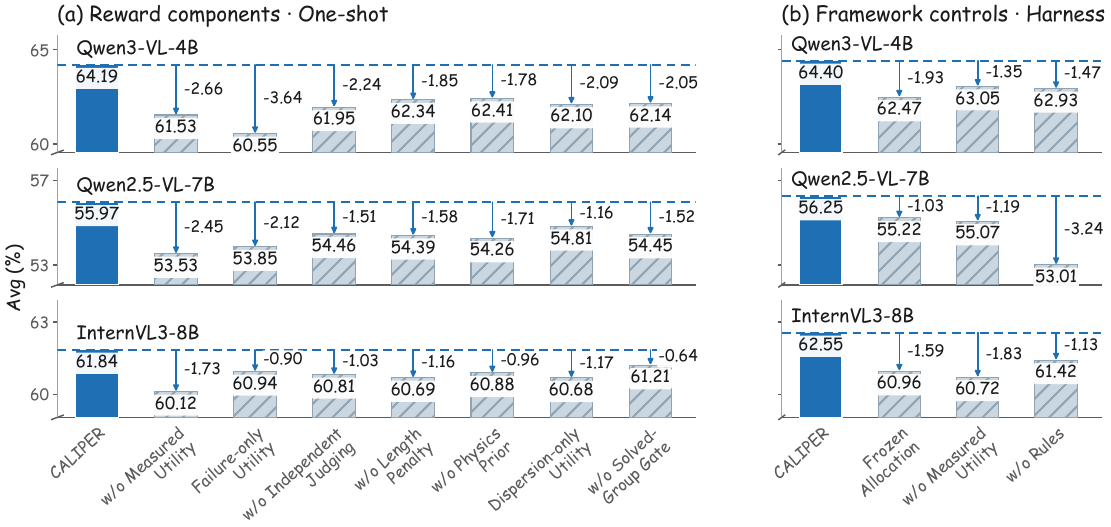}%
\vspace{-10pt}
\caption{Reward-component (a) and framework (b) ablations. Gaps are in percentage points.}
\label{fig:ablation}
\vspace{-4pt}
\end{figure}

\paragraph{Ablation Studies.}
Fig.~\ref{fig:ablation} shows consistent contributions from both adaptive reward allocation and harness control. Removing measured utility degrades all three backbones, with failure-only utility causing the largest drop on Qwen3-VL-4B ($3.64$ points), while freezing the learned allocation also consistently reduces performance. Removing harness rules produces further degradation across backbones, reaching $3.24$ points on Qwen2.5-VL-7B. Together, these results show that CALIPER benefits from both policy-dependent reward adaptation and closed-loop training control.

\begin{wrapfigure}{R}{0.62\textwidth}
\centering
\includegraphics[width=\linewidth]{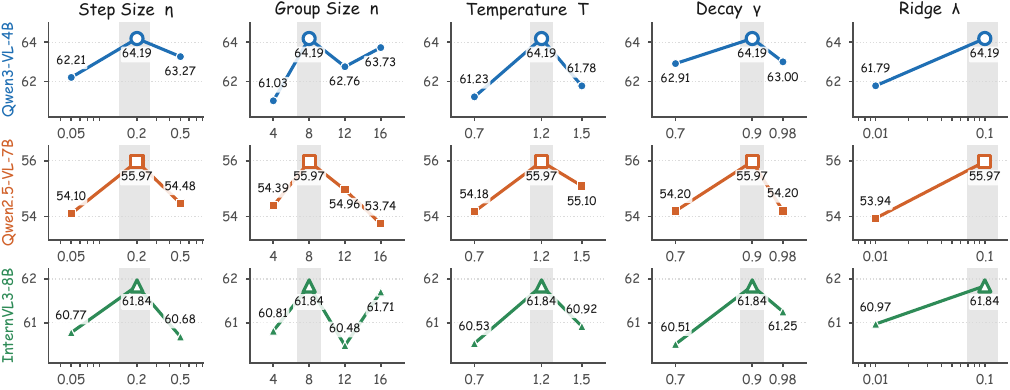}%
\vspace{-12pt}
\caption{Hyperparameter sensitivity.}
\label{fig:hparam}
\vspace{-12pt}
\end{wrapfigure}
\paragraph{Hyperparameter Sensitivity.}
Fig.~\ref{fig:hparam} shows that the shared default gives the highest mean on all three backbones. Group-size effects differ by backbone: Qwen3-VL-4B and InternVL3-8B recover at $n=16$ after a dip, whereas Qwen2.5-VL-7B declines beyond $n=8$. Non-default step size, temperature and decay also reduce performance, consistent with a balance between diverse rollouts and stable allocation updates. Increasing the ridge penalty from $0.01$ to $0.1$ improves the mean on all three backbones.

\begin{wrapfigure}{R}{0.6\textwidth}
\centering
\vspace{-12pt}
\resizebox{\linewidth}{!}{%
\begin{minipage}{\textwidth}
\begin{casebar}{Comparison of four reward formulations}
\noindent
\begin{minipage}[t]{0.175\linewidth}\vspace{0pt}\centering
\includegraphics[width=\linewidth]{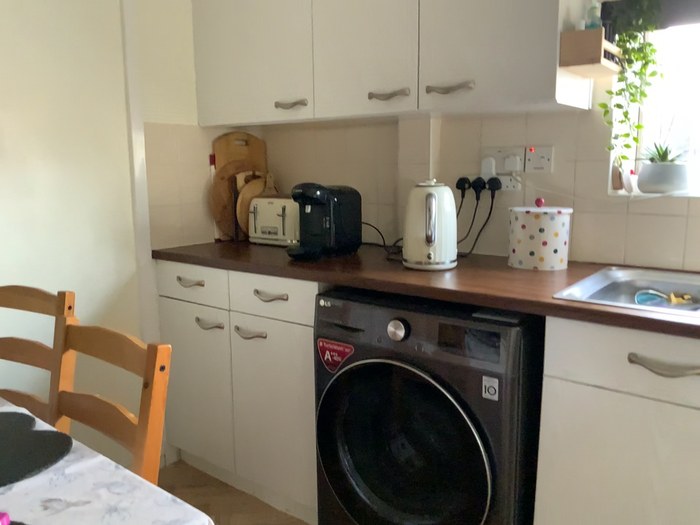}
\end{minipage}\hfill
\begin{minipage}[t]{0.805\linewidth}\vspace{0pt}
{\scriptsize\textbf{Qwen3-VL-4B, Omni3D-Beach.}~Which object is closer to the camera, the washing machine or the rightmost chair?}\par\vspace{1pt}
\stage{stageone}{Perception, the same in every run}
{\scriptsize\emph{``a washing machine on the right side of the counter, and a wooden chair\,\ldots{}\,on the left side, near the table''}}\par\vspace{2pt}
\stage{stagethree}{Inference}
\begin{tabular}{@{}l@{\hspace{3pt}}l@{\hspace{3pt}}p{0.80\linewidth}@{}}
{\scriptsize Base} & {\scriptsize \textcolor{lossred}{$\times$}} & {\scriptsize\raggedright \emph{``The chair is positioned further away from the camera\,\ldots{}\,The washing machine is positioned closer to the camera, as it is \hlc{in the foreground}''} $\rightarrow$ washing machine} \\
{\scriptsize Outcome} & {\scriptsize \textcolor{lossred}{$\times$}} & {\scriptsize\raggedright \emph{``The washing machine is positioned closer to the viewer than the chair, as it is \hlc{in the foreground}''} $\rightarrow$ washing machine} \\
{\scriptsize Dense} & {\scriptsize \textcolor{lossred}{$\times$}} & {\scriptsize\raggedright \emph{``The washing machine is positioned closer to the camera than the chair, as it is \hlc{in the foreground}''} $\rightarrow$ washing machine} \\
{\scriptsize \bfseries CALIPER} & {\scriptsize \textcolor{okgreen}{$\checkmark$}} & {\scriptsize\raggedright \emph{``The rightmost chair\,\ldots{}\,appears to be \hlc{in the foreground}. The washing machine is on the right side of the counter, slightly behind the chair''} $\rightarrow$ \hla{rightmost chair}} \\
\end{tabular}\par\vspace{1pt}
{\scriptsize\color{caserule}Every run places the two objects the same way and they disagree on which is nearer. Ground-truth answer: rightmost chair.}
\end{minipage}
\hrule\vspace{3.5pt}
\noindent
\begin{minipage}[t]{0.175\linewidth}\vspace{0pt}\centering
\includegraphics[width=\linewidth]{figures/cf_shampoo.jpg}
\end{minipage}\hfill
\begin{minipage}[t]{0.805\linewidth}\vspace{0pt}
{\scriptsize\textbf{Qwen3-VL-4B, RoboSpatial-Home.}~Can the shampoo fit in front of the floor mat? Answer yes or no.}\par\vspace{1pt}
\stage{stageone}{Perception, objects identified in every run}
{\scriptsize\emph{``There is a white floor mat on the dark wooden floor, positioned in front of the bathtub\,\ldots{}\,A white bottle of shampoo (or similar liquid soap) is placed on the edge of the bathtub''}}\par\vspace{2pt}
\stage{stagethree}{Decomposition and inference}
\begin{tabular}{@{}l@{\hspace{3pt}}l@{\hspace{3pt}}p{0.80\linewidth}@{}}
{\scriptsize Base} & {\scriptsize \textcolor{lossred}{$\times$}} & {\scriptsize\raggedright \emph{``The shampoo is positioned to the side of the tub, not in front of the mat\,\ldots{}\,the shampoo cannot fit \hlc{in front of the floor mat}''} $\rightarrow$ no} \\
{\scriptsize Outcome} & {\scriptsize \textcolor{lossred}{$\times$}} & {\scriptsize\raggedright \emph{``The shampoo bottle is positioned on the tub's edge, which is at a higher level than the floor mat\,\ldots{}\,the shampoo bottle is not physically in front of the mat''} $\rightarrow$ no} \\
{\scriptsize Dense} & {\scriptsize \textcolor{lossred}{$\times$}} & {\scriptsize\raggedright \emph{``The shampoo bottle is on the edge of the bathtub, which is behind the floor mat\,\ldots{}\,the shampoo bottle is not \hlc{in front of the floor mat}''} $\rightarrow$ no} \\
{\scriptsize \bfseries CALIPER} & {\scriptsize \textcolor{okgreen}{$\checkmark$}} & {\scriptsize\raggedright \emph{``This implies asking if the shampoo bottle can be placed in the space that is currently occupied by the floor mat or in front of it\,\ldots{}\,the shampoo bottle can be moved to the space \hlc{in front of the floor mat}''} $\rightarrow$ \hla{yes}} \\
\end{tabular}\par\vspace{1pt}
{\scriptsize\color{caserule}The baselines describe the current location, while CALIPER evaluates possible placement. Ground-truth answer: Yes.}
\end{minipage}
\end{casebar}
\end{minipage}%
}
\vspace{-8pt}
\caption{Qualitative effects of adaptive stage supervision.}
\label{fig:cases}
\vspace{-24pt}
\end{wrapfigure}

\paragraph{Case Study.}
Fig.~\ref{fig:cases} shows two failures to translate recognized objects into spatial answers. CALIPER resolves the camera-depth relation in the first item and distinguishes possible placement from current location in the second. Both cases illustrate gains in inference or decomposition despite similar object recognition. Appendix~\ref{app:cases} provides cross-backbone examples and a regression.

\begin{figure}[tbh]
\centering
\includegraphics[width=.98\linewidth]{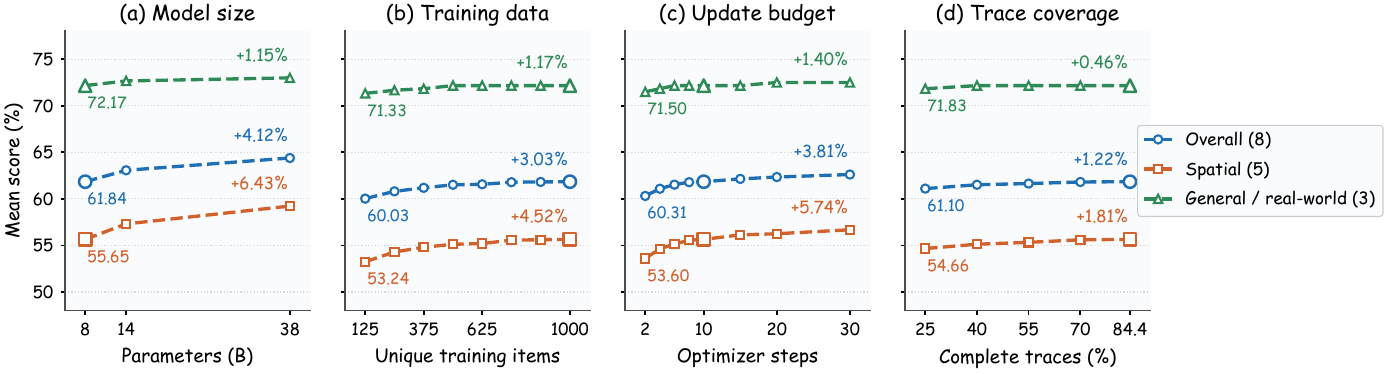}%
\vspace{-6pt}
\caption{Scaling with model size, training data, update budget and trace coverage.}
\vspace{-4pt}
\label{fig:scaling}
\end{figure}

\paragraph{Scaling Behaviour.}
Fig.~\ref{fig:scaling} shows diminishing gains with model capacity, unique training items, optimizer steps and complete-trace coverage. The spatial group gains more than the general and real-world group across these settings. The fixed-model panels vary data exposure, optimization and annotation coverage, while the model-size panel compares members of the InternVL3 family. Appendix~\ref{app:scaling} gives the corresponding settings.

\section{Conclusion}
This paper proposes CALIPER, an adaptive process-reward framework that allocates spatial-reasoning supervision according to policy-dependent stage utility. CALIPER estimates utility from answer-discriminative rollout groups and applies the resulting stage weights to all groups, including those with uniformly incorrect answers. A closed-loop harness monitors estimator reliability and validation performance to guide training decisions. Extensive experiments across eight benchmarks, three backbones and three optimizers show gains over matched fixed weights.

\subsection*{AI use statement}
In this work, we used AI tools for manuscript preparation, including language editing, selected icons in Fig.~\ref{fig:intro}, and code revision assistance. We also used a vision-language model to synthesize the initial SpatialTrace reasoning traces, which were subsequently refined by a text-only model. The authors take responsibility for the final manuscript, figures, dataset and code, including AI-assisted content. These tools did not independently determine the methods or technical conclusions.

\subsection*{Reproducibility statement}
Section~\ref{sec:data} and Appendix~\ref{app:data} describe SpatialTrace sources, sampling and reference annotation. Section~\ref{sec:method} specifies the reward, utility estimator and allocation update, with pseudocode in Algorithm~\ref{alg:caliper}. Appendix~\ref{app:impl} records training protocols, hyperparameters, evaluation metrics and reward-scoring prompts. Appendix~\ref{app:theory} provides the assumptions and proof of Proposition~\ref{prop:stage-contrast}, while Appendix~\ref{app:tables} details experimental controls and scaling settings.

\bibliographystyle{iclr2027_conference}
\bibliography{iclr2027_conference}

\clearpage
\appendix

\paragraph{Roadmap}
Appendix~\ref{app:data} details SpatialTrace construction. Appendix~\ref{app:impl} specifies implementation and evaluation protocols. Appendix~\ref{app:theory} proves Proposition~\ref{prop:stage-contrast}. Appendix~\ref{app:tables} provides detailed experiments and scaling settings, and Appendix~\ref{app:cases} presents additional qualitative examples.

\section{Dataset Construction Details}
\label{app:data}

\paragraph{Sources and Sampling.}
SpatialTrace combines ten single-image question sources grouped by reasoning demand in Table~\ref{tab:corpus}. The 3D family comprises 3DSRBench \citep{ma2024threedsrbench}, Omni3D-Bench \citep{brazil2023omni3d} and SAT-Spatial-VQA-5k \citep{ray2025satdynamicspatialaptitude}. General multimodal questions come from MMBench \citep{liu2023mmbench}, MME-RealWorld-Lite \citep{zhang2024mmerealworld} and MMStar \citep{chen2024mmstar}. Region-level spatial questions come from RoboSpatial-Home \citep{song2025robospatial}, SpatialEval \citep{wang2024spatialeval} and SpatialRGPT-Bench \citep{cheng2024spatialrgpt}, while SeePhys \citep{xiang2025seephys} adds physical reasoning. Records with empty question or answer fields or Chinese text were removed to match the English training prompt. Fixed quotas assign $310$ items to each core family and $250$ to physical reasoning. Within each core family, the largest-remainder method allocates items in proportion to filtered source sizes. Sampling uses seed $42$, followed by a source- and family-stratified split of $1{,}000$ training and $180$ validation items. Each $200$-item evaluation subset in Section~\ref{sec:exp} excludes every image in these splits, so SpatialTrace shares source benchmarks with the evaluation but no evaluation images.

\begin{table}[tbh]
\centering
\small
\setlength{\tabcolsep}{4pt}
\caption{SpatialTrace source families and reasoning demands.}
\label{tab:corpus}

\begin{tabular}{@{}>{\raggedright\arraybackslash}p{0.22\linewidth}>{\raggedright\arraybackslash}p{0.31\linewidth}c>{\raggedright\arraybackslash}p{0.29\linewidth}@{}}
\toprule
Family & Source datasets & Items & Primary reasoning demand \\
\midrule
3D metric & 3DSRBench, Omni3D-Bench, SAT-Spatial-VQA-5k & 310 & Depth, orientation and metric relations \\
\addlinespace[2pt]
General multimodal & MMBench, MME-RealWorld-Lite, MMStar & 310 & Object grounding and open-domain relations \\
\addlinespace[2pt]
Region-level spatial & RoboSpatial-Home, SpatialEval, SpatialRGPT-Bench & 310 & Placement and local spatial relations \\
\addlinespace[2pt]
Physical reasoning & SeePhys & 250 & Physical quantities and plausibility \\
\midrule
Total & Ten source datasets & 1,180 & Single-image questions with verifiable answers \\
\bottomrule
\end{tabular}
\end{table}

\paragraph{Reference Annotation and Coverage.}
Qwen2.5-VL-32B \citep{bai2025qwen25vltechnicalreport}, served locally through Ollama, generates visual observations, question decomposition and inference from each image, question and reference answer. The annotation prompt requests verifiable evidence and explicit acknowledgment of ambiguity. Decomposition identifies detection, counting and spatial subproblems, while inference connects observations to an answer. A text-only editing pass preserves visual observations, clarifies inferences and flags unsupported estimates or evidence--label conflicts. The three resulting fields form the stage references in Eq.~\ref{eq:reward}. Reflection and verification record annotation concerns without adding reward stages. Complete three-stage records cover $844$ training and $152$ validation items. The remaining $156$ training and $28$ validation items retain answer supervision and available reasoning. Policy training uses all $1{,}000$ training items, while utility estimation uses records with complete stage references.

\paragraph{Human Validation.}
Three unpaid members of the research group independently assess $100$ randomly sampled SpatialTrace items using the rubric in Table~\ref{tab:rubric}. They judge question validity and answer alignment from the image, question and reference answer before reading the stage references. Items without complete stage references are rated only on available stages. Table~\ref{tab:human_validation} reports positive rater-item judgments and Krippendorff's nominal $\alpha$ on the original response categories, which counts disagreement between adjacent ordinal levels as complete disagreement. Positive denotes a valid question, faithful perception, adequate decomposition, inference rated at least four, or agreement with the recorded answer alignment, respectively. Agreement is weakest for decomposition, where judging the completeness of a subproblem set requires the most interpretation.

\begin{table}[tbh]
\centering
\newcommand{\rubrictag}[1]{{\color{black!45}\textsc{#1}}}
\newcommand{\rubricline}[2]{\par\noindent\hangindent=2.6em\hangafter=1\rubrictag{#1}\enspace#2\par}
\newcommand{\rubricitem}[6]{%
  \noindent{\color{#1}\bfseries\scshape #2}\hfill{\color{black!45}\footnotesize #3}\par\vspace{2pt}
  \noindent{\itshape ``#4''}\par\vspace{2pt}
  \rubricline{positive}{#5}\vspace{1pt}
  \rubricline{failure}{#6}}
\caption{Rubric used by the SpatialTrace raters.}
\label{tab:rubric}
\small
\arrayrulecolor{black}
\begin{tabular}{@{}p{\linewidth}@{}}
\toprule
\rubricitem{caserule}{Question Validity}{item, binary}
  {Given the image and the question, can an attentive reader determine a single answer from visible evidence?}
  {The image contains the objects and relations the question asks about, and one answer follows from them.}
  {The evidence is occluded or outside the frame, the reference frame is unspecified, or several answers remain defensible.}\\
\arrayrulecolor{black!15}\midrule
\rubricitem{stageone}{Perception Fidelity}{perception, three-point ordinal}
  {Does each observation report objects, attributes or positions that are visible in the image and needed for the question?}
  {\emph{Faithful} reports only visible content and covers the evidence the answer depends on.}
  {\emph{Partial} is correct but omits one needed object or relation. \emph{Unfaithful} describes absent objects, misreads a position or orientation, or states the answer in place of evidence.}\\
\midrule
\rubricitem{stagetwo}{Decomposition Adequacy}{decomposition, three-point ordinal}
  {Would answering these subquestions in order suffice to answer the original question?}
  {\emph{Adequate} covers every object, quantity and reference frame the answer depends on.}
  {\emph{Partial} misses one of them. \emph{Inadequate} restates the question, targets irrelevant objects, or answers the subquestions itself.}\\
\midrule
\rubricitem{stagethree}{Inference Soundness}{inference, five-point Likert}
  {How well does the inference derive the reference answer from the recorded observations?}
  {\emph{5} every step follows from the observations and reaches the reference answer, \emph{4} it is sound apart from one minor gap.}
  {\emph{3} it reaches the answer through one weakly supported step, \emph{2} it relies on facts absent from the perception stage or on unjustified numerical precision, \emph{1} it contradicts the observations or reaches a different answer.}\\
\midrule
\rubricitem{caseflag}{Alignment Agreement}{item, binary}
  {Judging from the image, is the reference answer supported at its stated precision?}
  {The judgment, made before the recorded answer alignment is shown, matches that record.}
  {The rater finds an unsupported label or unjustified precision that the record accepts, or rejects a recorded concern.}\\
\arrayrulecolor{black}\bottomrule
\end{tabular}
\end{table}

\begin{table}[tbh]
\centering
\caption{Human assessment of SpatialTrace references.}
\label{tab:human_validation}
\small
\begin{tabular}{@{}lrrrc@{}}
\toprule
Criterion & Items & Positive ratings & Positive (\%) & $\alpha$ \\
\midrule
Question validity & 100 & 269/300 & 89.7 & 0.785 \\
Perception fidelity & 95 & 226/285 & 79.3 & 0.715 \\
Decomposition adequacy & 94 & 217/282 & 77.0 & 0.640 \\
Inference soundness & 95 & 212/285 & 74.4 & 0.808 \\
Alignment agreement & 100 & 255/300 & 85.0 & 0.765 \\
\bottomrule
\end{tabular}
\end{table}

\paragraph{Reference Traces.}
The reference examples distinguish visual evidence, decomposition and inference. The Omni3D example separates visible evidence from unsupported numerical precision, illustrating the complementary roles of reference coverage and physical plausibility.

\begin{casebox}{3DSRBench, 3D metric reasoning}
\noindent
\begin{minipage}[t]{0.20\linewidth}\vspace{0pt}\centering
\includegraphics[height=3.5cm]{figures/case_3dsr.jpg}
\end{minipage}\hfill
\begin{minipage}[t]{0.75\linewidth}\vspace{0pt}
\qa{Consider the real-world 3D locations and orientations of the objects. If I stand at the grey car's position facing where it is facing, is the fire hydrant on the left or right of me?}{B. \hla{on the right}}
\stage{stageone}{Perception}
\begin{tracelist}
\item A grey car is parked on the left side of the image, \hlc{facing towards the right}.
\item A red fire hydrant is located on the sidewalk near the intersection, \hlc{closer to the right side of the image}.
\end{tracelist}
\stage{stagetwo}{Decomposition}
\begin{tracelist}
\item What is the position and orientation of the grey car?
\item Where is the fire hydrant located relative to the car?
\item If standing at the car's position and facing its direction, which side is the fire hydrant on?
\end{tracelist}
\stage{stagethree}{Inference}
\begin{tracelist}
\item Step 1: The grey car is observed to be parked on the left side of the image and is \hlc{facing towards the right}. This establishes its orientation as directed to the right.
\item Step 2: The fire hydrant is located on the sidewalk near the intersection and is \hlc{closer to the right side of the image}. This places it spatially to the right of the car's orientation.
\item Step 3: If a person stands at the grey car's position and faces the direction the car is facing (towards the right), the fire hydrant, being positioned to the right of the car, would also be on the right side of the person.
\end{tracelist}
\end{minipage}
\end{casebox}

\begin{casebox}{Omni3D-Bench, 3D metric reasoning}
\noindent
\begin{minipage}[t]{0.26\linewidth}\vspace{0pt}\centering
\includegraphics[width=\linewidth]{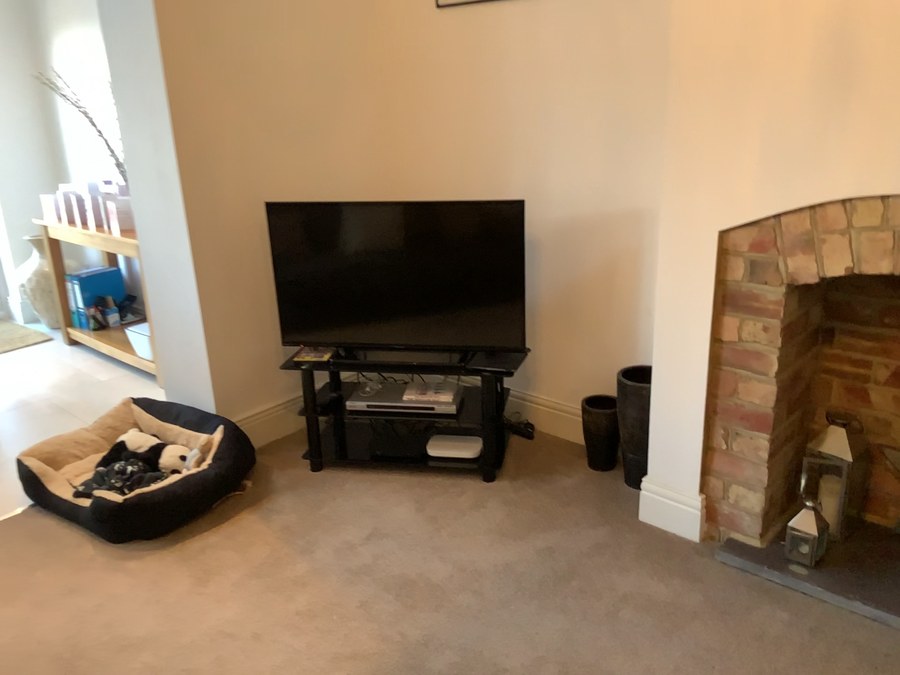}
\end{minipage}\hfill
\begin{minipage}[t]{0.69\linewidth}\vspace{0pt}
\qa{If the combined height of the TV and TV stand is 2.7m, how tall is the fireplace?}{\hla{2.989}}
\stage{stageone}{Perception}
\begin{tracelist}
\item The TV is mounted on a stand, and both are visible in the image.
\item The fireplace is visible on the right side of the image.
\item The TV and TV stand are positioned at the same height as the fireplace.
\item The height of the TV and stand is given as 2.7 meters.
\item The fireplace has a visible height that extends above the TV and stand.
\end{tracelist}
\stage{stagetwo}{Decomposition}
\begin{tracelist}
\item What is the height of the TV and stand? (Given as 2.7 meters)
\item How much taller is the fireplace compared to the TV and stand?
\item What is the total height of the fireplace?
\end{tracelist}
\stage{stagethree}{Inference}
\begin{tracelist}
\item Step 1: The height of the TV and stand is explicitly given as 2.7 meters.
\item Step 2: The fireplace is visually observed to extend above the top of the TV and stand. Based on the image, the additional height of the fireplace above the TV and stand appears to be slightly less than 0.3 meters.
\item Step 3: Using a coarse visual estimate, the additional height of the fireplace is \hlc{approximated as 0.289 meters}. This estimate is based on proportional comparison in the image, but no exact measurement tools are available.
\item Step 4: Adding the height of the TV and stand (2.7 meters) to the estimated additional height of the fireplace (0.289 meters) gives a total height of 2.989 meters for the fireplace.
\end{tracelist}
\stage{caseflag}{Answer alignment, recorded as not supported}
\begin{tracelist}
\item The stated precision of 2.989 meters exceeds what can be justified from the available visual evidence. The answer should be treated as a plausible estimate rather than an exact measurement.
\end{tracelist}
\end{minipage}
\end{casebox}

\section{Implementation Details}
\label{app:impl}
\label{app:hparams}

\begin{algorithm}[tbh]
\caption{CALIPER. Each harness round comprises $H$ steps of stage-adaptive GRPO.}
\label{alg:caliper}
\small
\begin{algorithmic}[1]
\Require SpatialTrace $\mathcal D$ with reference traces $\tau^\star$, policy $\pi_\theta$, stages $\mathcal S$
\Statex \textcolor{stageone}{\textbf{Settings:}} rounds $K$, steps per round $H$, patience $P$, step size $\eta$, offset $\epsilon$, decay $\gamma$, gate $N_{\min}$
\Ensure the policy of the best-validating round
\State $\alpha_0\gets$ uniform, statistics and $N_{\mathrm{eff}}\gets0$, $m^\star\gets-\infty$, $\Delta u_0\gets+\infty$
\State $\alpha\gets\alpha_0$
\For{round $k=1,\dots,K$}
  \Statex \hskip\algorithmicindent \textcolor{stageone}{\textit{Inner loop: stage-adaptive GRPO}}
  \For{step $t=1,\dots,H$}
    \State draw a batch from $\mathcal D$ and sample $n$ rollouts $\{o_i\}$ per question from $\pi_\theta$
    \State $\mathrm{cov}_s(o_i)\gets J(o_i,\tau^\star_s)$ for each $s\in\mathcal S$ \Comment{stages judged independently}
    \State score every rollout by Eq.~\ref{eq:reward} under the allocation $\alpha$ of the previous update
    \State form $A_i$ by Eq.~\ref{eq:grpo} and update $\theta$ by Eq.~\ref{eq:obj}
    \State $\mathcal M\gets\{g:0<\sum_{i\in g}\mathrm{acc}_i<n\}$ \Comment{answer-discriminative groups}
    \State decay statistics and $N_{\mathrm{eff}}$ by $\gamma$, then add complete-trace samples from $\mathcal M$
    \If{$N_{\mathrm{eff}}\ge N_{\min}$ and feature variation is nonzero}
      \State $u\gets$ Eq.~\ref{eq:utility}
      \If{$\max_s u_s>0$}
        \State $\alpha\gets$ Eq.~\ref{eq:alloc} \Comment{otherwise hold the allocation}
      \EndIf
    \EndIf
  \EndFor
  \Statex \hskip\algorithmicindent \textcolor{caseflag}{\textit{Outer loop: closed-loop harness}}
  \State $m_k\gets$ validation accuracy of $\theta$, and $\mathcal D_k\gets$ Eq.~\ref{eq:harness}
  \If{$m_k>m^\star$}
    \State $m^\star\gets m_k$ and retain the current checkpoint
  \EndIf
  \If{$\mathcal D_k=\textcolor{lossred}{\textsc{rollback}}$}
    \State restore the policy, allocation and rollout temperature of the best round
  \ElsIf{$\mathcal D_k=\textcolor{stagetwo}{\textsc{diversify}}$}
    \State raise the rollout temperature
  \ElsIf{$\mathcal D_k=\textcolor{caseflag}{\textsc{stop}}$}
    \State \textbf{break} \Comment{stages indistinguishable in utility}
  \EndIf
  \If{$m^\star$ has not improved for $P$ rounds}
    \State \textbf{break}
  \EndIf
\EndFor
\end{algorithmic}
\end{algorithm}

\paragraph{Framework and Optimization.}
EasyR1 implements GRPO on verl, with vLLM \citep{kwon2023efficient} for rollout generation and FSDP \citep{zhao2023pytorch} for policy updates. Rank-$64$ LoRA adapters with scaling $64$ update the language model's linear layers. The vision tower remains frozen. AdamW uses bfloat16, weight decay $10^{-2}$, gradient-norm clipping at $1.0$ and linear warmup over the first $5\%$ of training. Gradient checkpointing, unpadded packing and on-device states reduce memory overhead.

\paragraph{Rollouts and Batching.}
Each optimizer step processes $512$ training questions in four gradient updates of $128$. Rollouts use top-$p=1.0$, and prompts and responses are each limited to $3072$ tokens. Overlength prompts are excluded, and images are rescaled to $2.6{\times}10^{5}$--$4.2{\times}10^{6}$ pixels. GRPO uses a KL penalty of $10^{-2}$ against the frozen reference policy. DAPO and GSPO disable this penalty and use the optimizer-specific settings in Table~\ref{tab:hparams}. Data and reward definitions remain fixed across optimizers.

\paragraph{Protocols.}
One-shot training evaluates the final checkpoint after ten optimizer steps. Harness training uses at most six rounds of five steps, with early stopping after two rounds without validation improvement. Each round is evaluated on the $180$ validation items using greedy decoding, and the best validation checkpoint is selected for benchmark evaluation. Eq.~\ref{eq:harness} combines validation performance with allocation drift, group diversity and utility separation. Rollback requires a decline exceeding both the relative tolerance and twice the binomial standard error. Checkpoints retain the policy, allocation and rollout temperature.

\paragraph{Evaluation.}
Adapters are merged into the backbone for greedy decoding with a $2048$-token response budget. All methods use the same three-stage prompt except direct prompting, which retains the answer format but omits reasoning. Scoring is deterministic and task-specific. Multiple-choice items use option matching. Omni3D combines exact or categorical matching with mean relative accuracy over ten error thresholds for numerical answers. RoboSpatial-Home averages compatibility, configuration and context scores. Its context implementation uses the convex hull of reference points with a nearest-point fallback instead of the official mask. The reported mean weights the eight benchmark scores equally, preserving their task-specific scales rather than pooling them as binary outcomes.

\begin{table}[tbh]
\centering
\caption{Default hyperparameters and optimizer variants. AdamW uses bfloat16. Clipping values specify lower / upper thresholds.}
\label{tab:hparams}
\resizebox{\linewidth}{!}{%
\begin{tabular}[t]{@{}lcc@{}}
\toprule
\textbf{Hyperparameter} & \textbf{Symbol} & \textbf{Value} \\
\midrule
\multicolumn{3}{l}{\textit{Policy optimization}} \\
LoRA rank / scaling & & $64$ / $64$ \\
Rollouts per question & $n$ & $8$ \\
Rollout temperature & $T$ & $1.2$ \\
Questions / optimizer step & & $512$ \\
Gradient updates / step & & $4$ \\
AdamW learning rate & & $1{\times}10^{-5}$ \\
Maximum response length & & $3072$ \\
\midrule
\multicolumn{3}{l}{\textit{Process reward} (Eq.~\ref{eq:reward})} \\
Outcome weight & $w_a$ & $0.60$ \\
Process weight & $w_m$ & $0.25$ \\
Physics weight & $w_p$ & $0.08$ \\
Length weight & $w_\ell$ & $0.10$ \\
Length target & $L_0$ & $1500$ \\
Length cap & $L_{\max}$ & $3072$ \\
\midrule
\multicolumn{3}{l}{\textit{Utility estimator} (Eq.~\ref{eq:utility})} \\
\shortstack[l]{Ridge penalty /\\mean diagonal} & $\lambda$ & $0.1$ \\
Statistics decay & $\gamma$ & $0.9$ \\
Identifiability threshold & $N_{\min}$ & $24$ \\
\bottomrule
\end{tabular}\qquad
\begin{tabular}[t]{@{}lcc@{}}
\toprule
\textbf{Hyperparameter} & \textbf{Symbol} & \textbf{Value} \\
\midrule
\multicolumn{3}{l}{\textit{Allocation update} (Eq.~\ref{eq:alloc})} \\
Step size & $\eta$ & $0.2$ \\
Additive target offset & $\epsilon$ & $0.15$ \\
Initial allocation & $\alpha_0$ & uniform \\
\midrule
\multicolumn{3}{l}{\textit{Closed-loop harness} (Eq.~\ref{eq:harness})} \\
Steps per round & & $5$ \\
Maximum rounds / patience & & $6$ / $2$ \\
Regression tolerance & $\rho_{\mathrm r}$ & $0.02$ \\
Allocation drift threshold & $\epsilon_\delta$ & $0.02$ \\
\shortstack[l]{Answer-discriminative\\group threshold} & $\nu_{\min}$ & $1$ \\
Flat-utility threshold & $\epsilon_u$ & $0.05$ \\
Temperature step & $\Delta_T$ & $0.15$ \\
Temperature cap & $T_{\max}$ & $1.6$ \\
Round-selection items & & $180$ items \\
\midrule
\multicolumn{3}{l}{\textit{Optimizer variants} (Table~\ref{tab:algo})} \\
DAPO clipping range & & $0.2$ / $0.28$ \\
DAPO online filtering & & enabled \\
GSPO clipping range & & \shortstack{$3{\times}10^{-4}$ /\\$4{\times}10^{-4}$} \\
GSPO importance ratio & & sequence level \\
DAPO and GSPO KL penalty & & disabled \\
\bottomrule
\end{tabular}%
}
\end{table}

\subsection{Reward Scoring and Baseline Details}
\label{app:prompts}
\paragraph{Reward Computation.}
The policy generates three numbered reasoning stages followed by an answer. DeepSeek-V4-Flash \citep{deepseekai2026deepseekv4highlyefficientmilliontoken}, at temperature $0$, scores all three stages in one call by comparing each stage with its corresponding reference. Independent grading prevents evidence from one stage from inflating another stage's score. For incomplete references, the scorer excludes missing stages and renormalizes the remaining allocation. The same judge evaluates free-form answers if option matching is inapplicable. The physics term checks units, magnitudes and plausibility in the answer region and is active only if at least one check applies. The length penalty increases linearly from zero at $1500$ tokens to one at the $3072$-token cap, scaling the weighted reward as in Eq.~\ref{eq:reward}.

\paragraph{Process Reward Model Scorer.}
The PRM baseline combines a LoRA adapter with a three-output regression head on the policy backbone. It is fitted to stage labels obtained by scoring the untrained policy's rollouts against SpatialTrace references. During policy training, its predictions replace reference-based stage scores in Eq.~\ref{eq:reward} and receive equal weights. CALIPER additionally accesses reference traces at scoring time. The frozen-allocation control preserves that access and the judge, isolating the contribution of weight adaptation.

\paragraph{Reward Prompts.}
The judge computes the stage-coverage terms in Eq.~\ref{eq:reward} from one call per rollout. Inputs contain up to $500$ question characters, $6000$ reasoning characters and ten reference steps per stage. Stage labels distinguish visual observations, question decomposition and reasoning. The split clause indicates a multi-stage reference, and the strict clause requires evidence to match its designated stage. The loose-judging ablation replaces only that requirement, allowing a rollout step to match any reference stage.

\begin{promptbox}[stagetwo]{Stage-Matching Judge}
You are checking whether a model's reasoning covers the steps of a reference solution.{split_clause}

Question: {question}

{stage_blocks}

Model's reasoning:
{model_reasoning}

For EACH stage, count how many of its numbered reference steps have a semantically equivalent counterpart in the model's reasoning (same idea, not necessarily same words).{strict_clause}

Return ONLY a JSON object: {"visual_covered": <int>, "decomp_covered": <int>, "reasoning_covered": <int>}
\end{promptbox}

\begin{promptbox}[stagetwo]{Independence Instruction, Default and Loose Variant}
strict: Judge each stage independently: count a reference step as covered ONLY if
        the model's reasoning explicitly states its content, not merely because
        content from another stage implies it.

loose:  A step in the model's reasoning may cover steps from any stage.
\end{promptbox}

\paragraph{Baseline Access and Evaluation Prompts.}
The outcome-only judge receives the question, reference answer and predicted answer. Reference-assisted grading adds up to $800$ reference-trace characters. The PRM receives up to $400$ question characters and $2500$ reasoning characters without the reference trace. The frozen-allocation control holds reference access fixed to isolate weight adaptation. Trained methods and untrained chain-of-thought prompting share the three-stage response format. Direct prompting omits the reasoning request while retaining the answer format.

\section{Theoretical Analysis}
\label{app:theory}

\paragraph{Assumptions and Notation.}
Fix a policy and the conditional distributions of complete-reference rollout groups. Let $d=|\mathcal S|\ge2$, $x\in[0,1]^d$ collect stage-coverage scores, and $y\in\{0,1\}$ denote answer correctness. Within each group $g$, define $\tilde x=x-\bar x_g$ and $\tilde y=y-\bar y_g$ using its rollout means. Let $\mathcal M$ denote answer-discriminative groups and let $\mathcal M^c$ denote the remaining, outcome-degenerate groups, whose answers are either all correct or all wrong. The fraction of rollout mass in $\mathcal M$ is $p\in(0,1]$. Changing $p$ leaves both conditional distributions fixed. Define the conditional moments
\begin{equation}
\Sigma=\mathbb E[\tilde x\tilde x^\top\mid\mathcal M],\qquad
D=\mathbb E[\tilde x\tilde x^\top\mid\mathcal M^c],\qquad
c=\mathbb E[\tilde x\tilde y\mid\mathcal M].
\label{eq:theory-moments}
\end{equation}
These are exact expected moments or exact weighted empirical moments. Assume $\operatorname{tr}\Sigma>0$, $\operatorname{tr}D>0$, and fixed $\lambda,\epsilon>0$. Full rank is unnecessary. For a covariance matrix $H$, define
\begin{equation}
R(H)=H+\frac\lambda d\operatorname{tr}(H)I_d,\qquad
\mathcal T_\epsilon(u)=\frac{[u]_++\epsilon\mathbf1}{\|[u]_+\|_1+d\epsilon},\qquad
\alpha_{\mathrm{unif}}=\frac{\mathbf1}{d},
\label{eq:theory-maps}
\end{equation}
where $I_d$ is the identity, $\mathbf1$ is the all-ones vector, and $[u]_+$ is the componentwise positive part. Thus $R$ uses the trace-scaled penalty in Eq.~\ref{eq:utility}, and $\mathcal T_\epsilon$ is the target before interpolation in Eq.~\ref{eq:alloc}.

\paragraph{Formal Statement.}
Under these assumptions, the mixed-group and pooled estimators satisfy
\begin{equation}
u_{\mathrm M}=R(\Sigma)^{-1}c,\qquad
u_p=R\bigl(p\Sigma+(1-p)D\bigr)^{-1}pc.
\label{eq:theory-estimators}
\end{equation}
The utility $u_{\mathrm M}$ and target $\mathcal T_\epsilon(u_{\mathrm M})$ are independent of $p$. For $0<p<1$, the pooled target satisfies Eq.~\ref{eq:contrast-bound} and converges to $\alpha_{\mathrm{unif}}$ as $p\downarrow0$. Consequently, the limiting targets differ whenever the mixed-group target is nonuniform. This assertion concerns the computed target, before the algorithm's update gates.

For the normalized-advantage claim, take $d=3$, $v=(1,-1,0)^\top$, and any $a,b>0$ such that $a+b\le1/2$. Each group contains the four equally represented sign pairs $(z,\xi)\in\{-1,1\}^2$, repeated $m$ times for any integer $m\ge1$, so its size is $n=4m$. Define
\begin{equation}
x=\tfrac12\mathbf1+azv+b\xi\mathbf1,\qquad
y=\begin{cases}(z+1)/2 & \text{in mixed groups},\\ 0 & \text{in outcome-degenerate groups}.
\end{cases}
\label{eq:theory-family}
\end{equation}
The conditional feature distributions are identical, while only mixed groups associate the contrast $z$ with correctness. Let $\eta\in(0,1]$, initialize both allocations at $\alpha_{\mathrm{unif}}$, and apply $j\ge1$ eligible updates with fixed moments. Eligibility means that the effective-sample and feature-variation gates admit every update. Define
\begin{equation}
\begin{aligned}
r_\lambda&=\frac\lambda3(2a^2+3b^2),&
h&=\frac{a}{2(2a^2+r_\lambda)},&
\kappa_j&=1-(1-\eta)^j,\\
\delta_{\mathrm M,j}&=\kappa_j\frac{h}{h+3\epsilon},&
\delta_{p,j}&=\kappa_j\frac{ph}{ph+3\epsilon}.&&
\end{aligned}
\label{eq:theory-contrasts}
\end{equation}
On an all-wrong group from this family, assume a common physics score, a common positive length multiplier, and $w_m>0$. The advantages in Eq.~\ref{eq:grpo}, using the group standard deviation with divisor $n$, are
\begin{equation}
A_{\mathrm M,j}=\frac{a\delta_{\mathrm M,j}z+b\xi}{\sqrt{a^2\delta_{\mathrm M,j}^2+b^2}},\qquad
A_{p,j}=\frac{a\delta_{p,j}z+b\xi}{\sqrt{a^2\delta_{p,j}^2+b^2}}.
\label{eq:theory-advantages}
\end{equation}
For every $0<p<1$, their uniform within-group averages obey
\begin{equation}
0<\mathbb E[A_{p,j}z]<\mathbb E[A_{\mathrm M,j}z]
=\frac{a\delta_{\mathrm M,j}}{\sqrt{a^2\delta_{\mathrm M,j}^2+b^2}},\qquad
\lim_{p\downarrow0}\mathbb E[A_{p,j}z]=0.
\label{eq:theory-separation}
\end{equation}
The mixed-group association remains positive and independent of $p$. Here $z$ identifies the same stage-coverage contrast across both group types. It predicts correctness in mixed groups, while all evaluation-group answers in this construction are wrong.

\paragraph{Proof.}
Within any outcome-degenerate group, $y=\bar y_g$ for every rollout. Hence $\tilde y=0$ and $\mathbb E[\tilde x\tilde y\mid\mathcal M^c]=0$. The law of total expectation gives
\[
G_p=\mathbb E[\tilde x\tilde x^\top]=p\Sigma+(1-p)D,\qquad
\mathbb E[\tilde x\tilde y]=pc.
\]
Thus these groups contribute feature covariance but no feature--outcome cross-moment. For either conditional distribution and any vector $w$, the quadratic form of its covariance is $\mathbb E[(w^\top\tilde x)^2]\ge0$. Both $\Sigma$ and $D$, and therefore $G_p$, are positive semidefinite.

For a feature covariance $H$ and cross-moment $t$, expanding the ridge objective gives
\[
\mathbb E[\tilde y^2]-2\beta^\top t+\beta^\top H\beta
+\frac\lambda d\operatorname{tr}(H)\beta^\top\beta.
\]
Its gradient is $2R(H)\beta-2t$. For every nonzero $w$ and positive-trace $H$ considered here,
\[
w^\top R(H)w\ge\frac\lambda d\operatorname{tr}(H)\|w\|_2^2>0.
\]
The objective is strictly convex, so its unique minimizer is $R(H)^{-1}t$. Substitution of $(H,t)=(\Sigma,c)$ and $(G_p,pc)$ proves Eq.~\ref{eq:theory-estimators}. Neither $\Sigma$ nor $c$ depends on $p$, establishing invariance of the mixed-group utility and target. For weighted empirical data, dividing the squared-error sum and the Gram-matrix trace by the total sample weight scales the entire objective by the same positive constant. The minimizer is unchanged, giving exactly these moment formulas for Eq.~\ref{eq:utility}.

For $0<p<1$, positive semidefiniteness and the trace penalty imply
\[
R(G_p)\succeq\frac{(1-p)\lambda\operatorname{tr}D}{d}I_d.
\]
The spectral decomposition of this real symmetric positive-definite matrix bounds the Euclidean operator norm of its inverse by $d/((1-p)\lambda\operatorname{tr}D)$. Applying this inverse to $pc$ yields
\[
\|u_p\|_2\le\frac{pd\|c\|_2}{(1-p)\lambda\operatorname{tr}D}.
\]
Let $w=[u_p]_+$ and $s=\|w\|_1$. Subtracting the uniform vector from the allocation target gives
\[
\mathcal T_\epsilon(u_p)-\alpha_{\mathrm{unif}}
=\frac{w-(s/d)\mathbf1}{s+d\epsilon}.
\]
Since $w$ is nonnegative, the triangle inequality bounds the numerator's $\ell_1$ norm by $s+s=2s$. Cauchy--Schwarz and $\|[u_p]_+\|_2\le\|u_p\|_2$ give $s\le\sqrt d\|u_p\|_2$. Consequently,
\[
\bigl\|\mathcal T_\epsilon(u_p)-\alpha_{\mathrm{unif}}\bigr\|_1
\le\frac{2s}{s+d\epsilon}
\le\frac{2\sqrt d}{d\epsilon}\|u_p\|_2
\le\frac{2\sqrt d\,p\|c\|_2}{(1-p)\lambda\epsilon\operatorname{tr}D}.
\]
The last expression tends to zero as $p\downarrow0$. The mixed-group target is fixed, so it remains separated from this limit whenever it is nonuniform. This proves the target bound and limiting separation.

In Eq.~\ref{eq:theory-family}, the first two coordinates satisfy $|azv_s+b\xi|\le a+b\le1/2$, and the third satisfies $|b\xi|=b\le1/2$. Every coverage score therefore lies in $[0,1]$, and the specified labels are binary. Balanced sign pairs satisfy
\[
\mathbb E[z]=\mathbb E[\xi]=\mathbb E[z\xi]=0,\qquad
\mathbb E[z^2]=\mathbb E[\xi^2]=1.
\]
The group feature mean is $\mathbf1/2$, giving $\tilde x=azv+b\xi\mathbf1$. In mixed groups, $\bar y_g=1/2$ and $\tilde y=z/2$. In all-wrong groups, $\tilde y=0$. Expanding the products and using the zero cross-moment of the signs gives
\[
\Sigma=D=a^2vv^\top+b^2\mathbf1\mathbf1^\top,\qquad
c=\frac a2v,\qquad
\operatorname{tr}\Sigma=2a^2+3b^2.
\]
Since $v^\top v=2$ and $\mathbf1^\top v=0$, we obtain
\[
R(\Sigma)v=(2a^2+r_\lambda)v.
\]
Both group types have the same covariance, so $G_p=\Sigma$. Eq.~\ref{eq:theory-estimators} then gives $u_{\mathrm M}=hv$ and $u_p=phv$. Their first coordinate is positive for $p>0$, and the feature variance is nonzero. Thus the positivity and feature-variation gates admit both estimates. Replicating balanced groups increases the mixed-sample mass without changing the conditional moments, making the assumed effective-sample eligibility compatible with this construction.

For either fixed target $\alpha_{\mathrm{tar}}$, Eq.~\ref{eq:alloc} becomes $\alpha_{\ell+1}=(1-\eta)\alpha_\ell+\eta\alpha_{\mathrm{tar}}$. Starting at $\alpha_0=\alpha_{\mathrm{unif}}$, this recursion yields
\[
\alpha_j=(1-\eta)^j\alpha_{\mathrm{unif}}+\bigl(1-(1-\eta)^j\bigr)\alpha_{\mathrm{tar}}.
\]
The formula holds at $j=0$. If it holds at $j$, substitution into the recursion makes the coefficient on $\alpha_{\mathrm{unif}}$ equal to $(1-\eta)^{j+1}$ and the coefficient on $\alpha_{\mathrm{tar}}$ equal to $(1-\eta)(1-(1-\eta)^j)+\eta=1-(1-\eta)^{j+1}$. Induction proves the formula for every $j\ge0$. Each allocation stays in the simplex as a convex combination of two simplex vectors. Since $[hv]_+=(h,0,0)^\top$,
\[
\mathcal T_\epsilon(hv)=\frac{(h+\epsilon,\epsilon,\epsilon)^\top}{h+3\epsilon},\qquad
v^\top\mathcal T_\epsilon(hv)=\frac{h}{h+3\epsilon}.
\]
Replacing $h$ by $ph$ gives the pooled target's contrast $ph/(ph+3\epsilon)$. Also $v^\top\alpha_{\mathrm{unif}}=0$. Taking the inner product of the recursion solution with $v$ proves $v^\top\alpha_{\mathrm M,j}=\delta_{\mathrm M,j}$ and $v^\top\alpha_{p,j}=\delta_{p,j}$ in Eq.~\ref{eq:theory-contrasts}.

On an all-wrong group, the answer reward is zero. Let $P$ be the common physics score and $L=1-w_\ell\mathrm{len}>0$ the common length multiplier. For either allocation $\alpha$, Eq.~\ref{eq:reward} and $\mathbf1^\top\alpha=1$ give
\[
r=L\bigl(w_m\alpha^\top x+w_pP\bigr),\qquad
\alpha^\top x=\tfrac12+a(v^\top\alpha)z+b\xi.
\]
Write $\delta=v^\top\alpha$. Averaging over the balanced group removes both sign terms, so
\[
\bar r=L(w_m/2+w_pP),\qquad
r-\bar r=Lw_m(a\delta z+b\xi).
\]
Squaring the centered reward and averaging gives
\[
\sigma_r^2=L^2w_m^2\bigl(a^2\delta^2\mathbb E[z^2]
+2a\delta b\mathbb E[z\xi]+b^2\mathbb E[\xi^2]\bigr)
=L^2w_m^2(a^2\delta^2+b^2).
\]
The assumptions $L,w_m,b>0$ ensure $\sigma_r>0$, so the nondegenerate branch of Eq.~\ref{eq:grpo} applies. Division by $\sigma_r$ cancels the positive common factor $Lw_m$ and gives Eq.~\ref{eq:theory-advantages}. Multiplying by $z$ and averaging gives
\[
\mathbb E[Az]=F(\delta),\qquad
F(t)=\frac{at}{\sqrt{a^2t^2+b^2}},\qquad
F'(t)=\frac{ab^2}{(a^2t^2+b^2)^{3/2}}>0.
\]
For $j\ge1$, $\kappa_j>0$. Since $h,\epsilon>0$ and $0<p<1$, cross-multiplication gives $ph/(ph+3\epsilon)<h/(h+3\epsilon)$, equivalently $3p h\epsilon<3h\epsilon$. Hence $0<\delta_{p,j}<\delta_{\mathrm M,j}$, with $\delta_{p,j}\to0$ as $p\downarrow0$. Strict monotonicity and continuity of $F$, together with $F(0)=0$, prove Eq.~\ref{eq:theory-separation}. The mixed-group association stays positive and independent of $p$, completing the proof. $\square$

\section{Detailed Experiments}
\label{app:tables}
Table~\ref{tab:ablation:full} shows task-specific sensitivity to utility proxies. On Qwen3-VL-4B, failure-only utility produces the weakest SAT result among the tested component variants despite retaining competitive performance on other benchmarks. Grading and reward constraints have different task-specific effects, indicating that they do not substitute for estimating which stages distinguish correct answers. This pattern supports task-level assessment of utility beyond aggregate coverage.

Table~\ref{tab:harness:full} shows backbone-dependent sensitivity to harness settings. Relaxing the mixed-sample threshold approaches CALIPER on Qwen3-VL-4B but widens the gap on Qwen2.5-VL-7B. Frozen allocation and disabled rules also produce distinct task-level changes, separating reward adaptation from training intervention. This pattern is consistent with complementary contributions from informative rollout groups and the harness response to their scarcity.

Table~\ref{tab:algo:full} shows that CALIPER's advantage over frozen allocation extends to DAPO and GSPO. The gain varies by task despite matched optimizer settings and reference access within each comparison. This variation supports utility-based allocation across policy objectives, with task-dependent gains.
\begin{table*}[tbh]
\centering
\renewcommand{\arraystretch}{1.15}
\scriptsize
\caption{Component ablations by benchmark.}
\label{tab:ablation:full}
\resizebox{\textwidth}{!}{%
\begin{tabular}{lrrrrrrrrr}
\toprule
\textbf{Method} & \textsc{3DSR} & \textsc{MMB} & \textsc{MME-RW} & \textsc{MMStar} & \textsc{Omni3D} & \textsc{RoboSp} & \textsc{SAT} & \textsc{SpatEval} & \textbf{Avg} \\
\midrule
\multicolumn{10}{l}{\itshape Qwen3-VL-4B} \\[1pt]
\rowcolor{ourrow}CALIPER (ours) & \underline{66.00} & \bfseries 93.50 & \bfseries 52.00 & \bfseries 69.00 & \bfseries 37.05 & \bfseries 46.95 & \bfseries 76.50 & 72.50 & \bfseries 64.19 \\
w/o Measured Utility & 61.50 & \underline{92.50} & 48.00 & \underline{67.50} & 32.05 & 44.69 & 73.50 & 72.50 & 61.53 \\
w/o Independent Judging & 59.50 & 92.00 & 50.00 & 66.00 & 31.55 & \underline{45.06} & \underline{76.00} & \underline{75.50} & 61.95 \\
w/o Solved-Group Gate & 62.50 & 91.00 & 49.50 & 67.00 & 35.00 & 43.13 & 74.00 & 75.00 & 62.14 \\
w/o Length Penalty & 64.00 & 91.00 & \underline{50.50} & 66.50 & 36.70 & 40.48 & 74.50 & 75.00 & 62.34 \\
w/o Physics Prior & \bfseries 67.00 & 92.00 & \underline{50.50} & 65.50 & 33.10 & 43.15 & 74.00 & 74.00 & \underline{62.41} \\
w/o Utility, Dispersion & 61.50 & \underline{92.50} & 49.00 & 67.00 & 32.75 & 43.56 & 75.50 & 75.00 & 62.10 \\
w/o Utility, Failure & 61.50 & 92.00 & \underline{50.50} & 67.00 & \underline{36.90} & 44.97 & 54.50 & \bfseries 77.00 & 60.55 \\
\addlinespace[2pt]\midrule
\multicolumn{10}{l}{\itshape Qwen2.5-VL-7B} \\[1pt]
\rowcolor{ourrow}CALIPER (ours) & \bfseries 54.50 & \underline{90.50} & \bfseries 50.50 & 63.50 & \bfseries 23.05 & 46.73 & 57.00 & \bfseries 62.00 & \bfseries 55.97 \\
w/o Measured Utility & 51.50 & 88.00 & 46.00 & 61.50 & \bfseries 23.05 & 45.66 & 56.00 & 56.50 & 53.53 \\
w/o Independent Judging & 51.50 & 90.00 & 48.50 & 61.00 & \underline{22.95} & 45.72 & \underline{58.50} & 57.50 & 54.46 \\
w/o Solved-Group Gate & 51.50 & 90.00 & 46.50 & \underline{64.50} & 22.50 & 45.09 & 56.00 & 59.50 & 54.45 \\
w/o Length Penalty & 51.00 & 88.00 & \underline{49.50} & 63.00 & 21.55 & \underline{47.07} & 55.00 & \underline{60.00} & 54.39 \\
w/o Physics Prior & \underline{52.00} & 89.50 & \underline{49.50} & 63.50 & 20.45 & 45.61 & 56.00 & 57.50 & 54.26 \\
w/o Utility, Dispersion & 51.00 & 90.00 & 49.00 & 63.00 & 21.85 & \bfseries 47.17 & \bfseries 59.00 & 57.50 & \underline{54.81} \\
w/o Utility, Failure & 49.50 & \bfseries 91.00 & 48.50 & \bfseries 65.50 & 21.25 & 43.58 & 55.00 & 56.50 & 53.85 \\
\addlinespace[2pt]\midrule
\multicolumn{10}{l}{\itshape InternVL3-8B} \\[1pt]
\rowcolor{ourrow}CALIPER (ours) & \bfseries 58.00 & \underline{89.00} & \underline{55.00} & \bfseries 72.50 & \bfseries 41.65 & 46.10 & 75.00 & \bfseries 57.50 & \bfseries 61.84 \\
w/o Measured Utility & 55.50 & 88.50 & 54.00 & 70.50 & 37.70 & 46.73 & 75.50 & 52.50 & 60.12 \\
w/o Independent Judging & 56.50 & \bfseries 89.50 & 54.00 & \underline{72.00} & 37.30 & 48.22 & 73.50 & 55.50 & 60.81 \\
w/o Solved-Group Gate & 55.00 & \underline{89.00} & \bfseries 56.00 & 71.00 & 36.90 & \bfseries 49.75 & 75.50 & \underline{56.50} & \underline{61.21} \\
w/o Length Penalty & 56.50 & 88.00 & 53.50 & 69.50 & 37.15 & 47.83 & \bfseries 77.00 & 56.00 & 60.69 \\
w/o Physics Prior & 56.50 & \underline{89.00} & \underline{55.00} & \bfseries 72.50 & 35.25 & 47.82 & 74.50 & \underline{56.50} & 60.88 \\
w/o Utility, Dispersion & 56.50 & 87.50 & 54.00 & 70.00 & 36.20 & \underline{48.71} & \underline{76.00} & \underline{56.50} & 60.68 \\
w/o Utility, Failure & \underline{57.50} & 88.00 & \underline{55.00} & 69.50 & \underline{39.80} & 45.73 & \underline{76.00} & 56.00 & 60.94 \\
\bottomrule
\end{tabular}}
\end{table*}

\begin{table*}[tbh]
\centering
\renewcommand{\arraystretch}{1.15}
\scriptsize
\caption{Harness ablations by benchmark.}
\label{tab:harness:full}
\resizebox{\textwidth}{!}{%
\begin{tabular}{lrrrrrrrrr}
\toprule
\textbf{Method} & \textsc{3DSR} & \textsc{MMB} & \textsc{MME-RW} & \textsc{MMStar} & \textsc{Omni3D} & \textsc{RoboSp} & \textsc{SAT} & \textsc{SpatEval} & \textbf{Avg} \\
\midrule
\multicolumn{10}{l}{\itshape Qwen3-VL-4B} \\[1pt]
Outcome Reward & 58.00 & 91.00 & 50.50 & 64.50 & 33.65 & 46.72 & 76.00 & 70.50 & 61.36 \\
Reference-Assisted & 61.50 & 92.00 & 50.00 & 65.00 & 32.55 & 48.03 & 75.00 & 71.50 & 61.95 \\
Frozen $\alpha$ & 64.50 & 94.00 & 50.50 & 65.50 & 32.70 & 44.09 & 75.50 & 73.00 & 62.47 \\
w/o Measured Utility & 62.50 & 93.00 & 50.00 & 69.00 & 35.20 & 48.19 & 73.50 & 73.00 & 63.05 \\
$\alpha$ EMA $=0.05$ & 60.00 & 92.50 & 52.50 & 65.50 & 35.85 & 49.58 & 77.00 & 73.50 & 63.30 \\
Min.\ Mixed Samples $=8$ & 65.00 & 93.00 & 50.50 & 68.50 & 34.15 & 46.46 & 76.50 & 79.00 & 64.14 \\
Rollout Temperature $=1.5$ & 60.50 & 91.50 & 49.00 & 70.50 & 32.35 & 48.61 & 76.50 & 79.00 & 63.49 \\
w/o Rules & 62.00 & 93.50 & 53.00 & 65.00 & 34.35 & 47.61 & 74.00 & 74.00 & 62.93 \\
\rowcolor{ourrow}CALIPER~(ours)  & 65.50 & 93.00 & 52.50 & 70.00 & 36.35 & 50.37 & 72.50 & 75.00 & 64.40 \\
\addlinespace[2pt]\midrule
\multicolumn{10}{l}{\itshape Qwen2.5-VL-7B} \\[1pt]
Outcome Reward & 55.50 & 89.50 & 49.50 & 65.50 & 24.05 & 43.06 & 62.50 & 55.50 & 55.64 \\
Reference-Assisted & 53.00 & 88.50 & 50.00 & 63.50 & 22.45 & 43.47 & 62.00 & 57.00 & 54.99 \\
Frozen $\alpha$ & 52.50 & 89.50 & 50.50 & 65.00 & 22.60 & 46.65 & 55.50 & 59.50 & 55.22 \\
w/o Measured Utility & 51.00 & 90.50 & 52.00 & 65.00 & 20.95 & 43.58 & 58.50 & 59.00 & 55.07 \\
$\alpha$ EMA $=0.05$ & 51.50 & 89.50 & 49.50 & 64.50 & 23.90 & 46.18 & 56.50 & 59.00 & 55.07 \\
Min.\ Mixed Samples $=8$ & 52.50 & 90.00 & 48.00 & 65.00 & 19.80 & 44.10 & 53.50 & 57.50 & 53.80 \\
Rollout Temperature $=1.5$ & 52.00 & 90.50 & 48.00 & 65.00 & 24.65 & 44.15 & 58.00 & 57.50 & 54.98 \\
w/o Rules & 49.00 & 87.00 & 46.50 & 63.00 & 21.40 & 45.66 & 55.50 & 56.00 & 53.01 \\
\rowcolor{ourrow}CALIPER~(ours)  & 53.50 & 90.50 & 49.50 & 65.50 & 24.75 & 48.27 & 55.50 & 62.50 & 56.25 \\
\addlinespace[2pt]\midrule
\multicolumn{10}{l}{\itshape InternVL3-8B} \\[1pt]
Outcome Reward & 55.50 & 87.50 & 54.50 & 66.50 & 37.00 & 43.75 & 74.00 & 57.50 & 59.53 \\
Reference-Assisted & 55.00 & 88.00 & 54.00 & 68.50 & 38.10 & 47.19 & 74.00 & 57.50 & 60.29 \\
Frozen $\alpha$ & 55.50 & 88.50 & 55.00 & 73.50 & 37.95 & 47.20 & 74.50 & 55.50 & 60.96 \\
w/o Measured Utility & 56.00 & 89.00 & 56.00 & 70.00 & 35.40 & 45.32 & 75.50 & 58.50 & 60.72 \\
$\alpha$ EMA $=0.05$ & 57.50 & 88.50 & 54.00 & 72.00 & 37.30 & 46.65 & 74.50 & 57.00 & 60.93 \\
Min.\ Mixed Samples $=8$ & 58.50 & 89.00 & 54.50 & 75.00 & 37.60 & 46.75 & 76.50 & 55.00 & 61.61 \\
w/o Rules & 57.50 & 88.50 & 54.50 & 69.50 & 38.60 & 50.24 & 75.50 & 57.00 & 61.42 \\
\rowcolor{ourrow}CALIPER~(ours)  & 57.50 & 89.00 & 55.50 & 71.00 & 40.25 & 52.14 & 75.00 & 60.00 & 62.55 \\
\bottomrule
\end{tabular}}
\end{table*}

\begin{table*}[tbh]
\centering
\renewcommand{\arraystretch}{1.15}
\scriptsize
\caption{Optimizer comparisons under harness training.}
\label{tab:algo:full}
\resizebox{\textwidth}{!}{%
\begin{tabular}{lrrrrrrrrr}
\toprule
\textbf{Method} & \textsc{3DSR} & \textsc{MMB} & \textsc{MME-RW} & \textsc{MMStar} & \textsc{Omni3D} & \textsc{RoboSp} & \textsc{SAT} & \textsc{SpatEval} & \textbf{Avg} \\
\midrule
\multicolumn{10}{l}{\itshape Qwen3-VL-4B} \\[1pt]
GRPO, Outcome Reward & 58.00 & 91.00 & 50.50 & 64.50 & 33.65 & 46.72 & 76.00 & 70.50 & 61.36 \\
GRPO, Reference-Assisted & 61.50 & 92.00 & 50.00 & 65.00 & 32.55 & 48.03 & 75.00 & 71.50 & 61.95 \\
GRPO, Frozen $\alpha$ & 64.50 & \bfseries 94.00 & 50.50 & 65.50 & 32.70 & 44.09 & 75.50 & 73.00 & 62.47 \\
GRPO, CALIPER~(ours)  & \bfseries 65.50 & 93.00 & \bfseries 52.50 & \bfseries 70.00 & \underline{36.35} & 50.37 & 72.50 & \underline{75.00} & \bfseries 64.40 \\
DAPO, Outcome Reward & 61.50 & \underline{93.50} & 50.50 & 65.50 & 34.45 & \underline{51.03} & \underline{76.50} & 73.50 & 63.31 \\
DAPO, Reference-Assisted & 62.50 & \bfseries 94.00 & \bfseries 52.50 & 66.50 & 31.70 & \bfseries 51.18 & 75.50 & 73.00 & \underline{63.36} \\
DAPO, Frozen $\alpha$ & 64.50 & 91.00 & 51.00 & \underline{69.00} & 33.90 & 43.11 & 73.50 & 74.50 & 62.56 \\
DAPO, CALIPER~(ours) & \bfseries 65.50 & 92.00 & 50.50 & 64.50 & 33.60 & 50.19 & 74.50 & \bfseries 76.00 & 63.35 \\
GSPO, Outcome Reward & 60.00 & 92.00 & \underline{52.00} & 66.00 & 32.45 & 45.03 & 74.00 & 70.00 & 61.43 \\
GSPO, Reference-Assisted & \underline{65.00} & 91.50 & 51.00 & 66.50 & 32.00 & 47.62 & \bfseries 77.00 & 73.00 & 62.95 \\
GSPO, Frozen $\alpha$ & \bfseries 65.50 & 92.00 & \underline{52.00} & 65.00 & \bfseries 37.65 & 44.59 & 74.50 & 71.00 & 62.78 \\
GSPO, CALIPER~(ours) & 64.00 & 91.50 & 50.00 & 67.00 & 35.05 & 50.15 & 75.00 & 72.50 & 63.15 \\
\addlinespace[2pt]\midrule
\multicolumn{10}{l}{\itshape Qwen2.5-VL-7B} \\[1pt]
GRPO, Outcome Reward & \bfseries 55.50 & 89.50 & 49.50 & \bfseries 65.50 & 24.05 & 43.06 & \underline{62.50} & 55.50 & 55.64 \\
GRPO, Reference-Assisted & 53.00 & 88.50 & 50.00 & 63.50 & 22.45 & 43.47 & 62.00 & 57.00 & 54.99 \\
GRPO, Frozen $\alpha$ & 52.50 & 89.50 & 50.50 & \underline{65.00} & 22.60 & 46.65 & 55.50 & 59.50 & 55.22 \\
GRPO, CALIPER~(ours)  & 53.50 & \bfseries 90.50 & 49.50 & \bfseries 65.50 & \underline{24.75} & \bfseries 48.27 & 55.50 & \bfseries 62.50 & 56.25 \\
DAPO, Outcome Reward & \bfseries 55.50 & 89.00 & \underline{53.00} & 62.00 & 23.35 & 45.09 & \bfseries 63.00 & 59.50 & \bfseries 56.31 \\
DAPO, Reference-Assisted & \underline{55.00} & 87.50 & \bfseries 53.50 & 63.50 & \bfseries 25.00 & 44.05 & 61.50 & \underline{60.00} & \underline{56.26} \\
DAPO, Frozen $\alpha$ & 49.50 & 86.50 & 47.00 & 63.00 & 24.35 & 44.67 & 55.50 & 56.50 & 53.38 \\
DAPO, CALIPER~(ours) & 51.00 & 85.00 & 48.00 & 61.00 & 23.55 & \underline{48.21} & 59.50 & 55.50 & 53.97 \\
GSPO, Outcome Reward & \bfseries 55.50 & 89.50 & 52.00 & 64.00 & 22.60 & 45.04 & 56.50 & 58.50 & 55.46 \\
GSPO, Reference-Assisted & \bfseries 55.50 & 86.50 & 48.50 & 62.00 & \underline{24.75} & 42.02 & 57.50 & 54.50 & 53.91 \\
GSPO, Frozen $\alpha$ & 52.00 & 87.00 & 48.00 & 60.50 & 18.85 & 43.68 & 56.00 & 56.00 & 52.75 \\
GSPO, CALIPER~(ours) & 51.00 & \underline{90.00} & 50.00 & 64.50 & 22.80 & 45.08 & 57.00 & 57.00 & 54.67 \\
\addlinespace[2pt]\midrule
\multicolumn{10}{l}{\itshape InternVL3-8B} \\[1pt]
GRPO, Outcome Reward & 55.50 & 87.50 & 54.50 & 66.50 & 37.00 & 43.75 & 74.00 & 57.50 & 59.53 \\
GRPO, Reference-Assisted & 55.00 & 88.00 & 54.00 & 68.50 & 38.10 & 47.19 & 74.00 & 57.50 & 60.29 \\
GRPO, Frozen $\alpha$ & 55.50 & 88.50 & 55.00 & \bfseries 73.50 & 37.95 & 47.20 & \underline{74.50} & 55.50 & 60.96 \\
GRPO, CALIPER~(ours)  & \underline{57.50} & \underline{89.00} & \underline{55.50} & 71.00 & \bfseries 40.25 & \underline{52.14} & \bfseries 75.00 & 60.00 & \bfseries 62.55 \\
DAPO, Outcome Reward & 54.50 & \underline{89.00} & 52.50 & 71.00 & 38.25 & 43.50 & 74.00 & \bfseries 68.00 & 61.34 \\
DAPO, Reference-Assisted & 55.00 & \bfseries 89.50 & 54.00 & 69.50 & 36.10 & 48.10 & 71.00 & \underline{66.50} & 61.21 \\
DAPO, Frozen $\alpha$ & \bfseries 59.00 & \bfseries 89.50 & \underline{55.50} & 70.50 & 34.30 & 49.38 & 73.00 & 57.50 & 61.09 \\
DAPO, CALIPER~(ours) & 57.00 & 88.50 & \bfseries 56.50 & \underline{73.00} & 39.15 & \bfseries 52.17 & \bfseries 75.00 & 58.50 & \underline{62.48} \\
GSPO, Outcome Reward & 56.00 & 88.00 & 52.00 & 70.00 & 39.10 & 46.67 & \underline{74.50} & 53.50 & 59.97 \\
GSPO, Reference-Assisted & \underline{57.50} & 88.00 & 51.50 & 71.00 & \underline{39.20} & 46.14 & 71.00 & 57.00 & 60.17 \\
GSPO, Frozen $\alpha$ & 57.00 & \underline{89.00} & 53.00 & 72.50 & 38.30 & 45.26 & 72.00 & 56.50 & 60.44 \\
GSPO, CALIPER~(ours) & \underline{57.50} & \underline{89.00} & 53.50 & 70.00 & 37.90 & 47.78 & 73.00 & 56.00 & 60.58 \\
\bottomrule
\end{tabular}}
\end{table*}

\subsection{Hyperparameter Sensitivity}
\begin{table}[tbh]
\centering\small
\renewcommand{\arraystretch}{1.16}
\setlength{\tabcolsep}{10pt}
\caption{One-shot sensitivity averaged over eight benchmarks.}
\label{tab:hparam:summary}
\begin{tabular}{@{}lrrr@{}}
\toprule
Configuration & Qwen3-VL-4B & Qwen2.5-VL-7B & InternVL3-8B \\
\midrule
\rowcolor{ourrow}Default configuration & 64.19 & 55.97 & 61.84 \\
$\eta=0.05$ & 62.21 & 54.10 & 60.77 \\
$\eta=0.5$ & 63.27 & 54.48 & 60.68 \\
$\lambda=0.01$ & 61.79 & 53.94 & 60.97 \\
$n=4$ & 61.03 & 54.39 & 60.81 \\
$n=12$ & 62.76 & 54.96 & 60.48 \\
$n=16$ & 63.73 & 53.74 & 61.71 \\
$T=0.7$ & 61.23 & 54.18 & 60.53 \\
$T=1.5$ & 61.78 & 55.10 & 60.92 \\
$\gamma=0.7$ & 62.91 & 54.20 & 60.51 \\
$\gamma=0.98$ & 63.00 & 54.20 & 61.25 \\
\bottomrule\end{tabular}\end{table}

Table~\ref{tab:hparam:summary} shows that the shared default yields the highest mean across the tested settings for each backbone. Larger rollout groups partly recover performance on Qwen3-VL-4B and InternVL3-8B but reduce it on Qwen2.5-VL-7B. The nonmonotonic response indicates that larger rollout groups alone do not consistently improve performance.

\subsection{Scaling Settings}
\label{app:scaling}
\paragraph{Model Size.}
Fig.~\ref{fig:scaling}a compares the 8B, 14B and 38B variants of InternVL3 \citep{zhu2025internvl3} with $1{,}000$ training items, rank-$64$ LoRA, ten update steps and eight rollouts per question. The 38B variant also changes its vision encoder, so this panel describes family-level capacity scaling.

\paragraph{Training Data.}
The data-size panel fixes InternVL3-8B and uses nested subsets that approximately preserve source proportions and complete-trace coverage. Each update samples $512$ questions with replacement, holding the update count fixed while varying exposure to unique items.

\paragraph{Update Budget.}
The update panel evaluates final checkpoints after 2, 4, 6, 8, 10, 15, 20 and 30 steps. All settings use $1{,}000$ training items, eight rollouts per question and $512$ questions per step, isolating the number of updates.

\paragraph{Trace Coverage.}
The annotation panel fixes ten updates and varies complete-trace coverage over $\{25,40,55,70,84.4\}\%$. Nested, source-stratified subsets retain all stages for selected complete traces. Other records retain answers and available reasoning, with missing stages masked. The fixed $1{,}000$-item training set isolates stage evidence for utility estimation.

\section{Qualitative Evidence}
\label{app:cases}
The left--right example documents an inference error despite an accurate visual description, separating evidence recognition from its use in the answer. In the Omni3D example, the InternVL3 fixed-weight trace identifies the nearer chair but selects the washing machine, whereas CALIPER follows the depth evidence. In the RoboSpatial placement example, CALIPER answers correctly on both backbones while the other rewards exhibit backbone-dependent errors.
\begin{figure}[tbh]
\begin{casebox}{RoboSpatial-Home, left--right regression}
\noindent
\begin{minipage}[t]{0.20\linewidth}\vspace{0pt}\centering
\includegraphics[width=\linewidth]{figures/cf_regression_speaker.jpeg}
\end{minipage}\hfill
\begin{minipage}[t]{0.76\linewidth}\vspace{0pt}
{\scriptsize Can the speaker fit left of the cup? Answer yes or no.}\par\vspace{2pt}
\stage{stageone}{Perception and inference}
\begin{tabular}{@{}l@{\hspace{3pt}}l@{\hspace{2pt}}p{0.66\linewidth}@{}}
{\scriptsize IVL3-8B Base} & {\scriptsize \textcolor{okgreen}{$\checkmark$}} & {\scriptsize\raggedright \emph{``The speaker is located to the left of the cup in the image.''} $\rightarrow$ \hla{yes}} \\
{\scriptsize IVL3-8B Outcome} & {\scriptsize \textcolor{okgreen}{$\checkmark$}} & {\scriptsize\raggedright \emph{``the speaker is already positioned to the left of the cup''} $\rightarrow$ \hla{yes}} \\
{\scriptsize IVL3-8B Dense} & {\scriptsize \textcolor{okgreen}{$\checkmark$}} & {\scriptsize\raggedright \emph{``the speaker is currently to the left of the cup''} $\rightarrow$ \hla{yes}} \\
{\scriptsize \bfseries IVL3-8B CALIPER} & {\scriptsize \textcolor{lossred}{$\times$}} & {\scriptsize\raggedright \emph{``The speaker is located to the left of the computer monitor, and the cup is placed to the right of the monitor on the desk.''} Yet it concludes \emph{``the speaker is not to the left of the cup''} $\rightarrow$ no} \\
\end{tabular}\par\vspace{1pt}
{\scriptsize\itshape\color{caserule}Ground-truth answer: Yes.}
\end{minipage}
\end{casebox}
\end{figure}

\begin{figure}[tbh]
\begin{casebox}{Omni3D-Bench, both backbones on one item}
\noindent
\begin{minipage}[t]{0.20\linewidth}\vspace{0pt}\centering
\includegraphics[width=\linewidth]{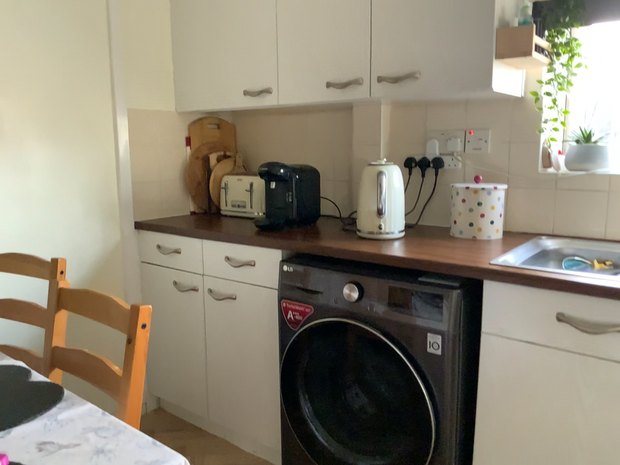}
\end{minipage}\hfill
\begin{minipage}[t]{0.76\linewidth}\vspace{0pt}
{\scriptsize Which object is closer to the camera: the washing machine or the rightmost chair?}\par\vspace{1.5pt}
\stage{stageone}{Perception}
{\scriptsize\emph{``The image shows a kitchen area with a washing machine on the right side of the counter, and a wooden chair (partially visible) on the left side, near the\,\ldots{}''}}\par\vspace{2pt}
\stage{stagethree}{Inference}
\begin{tabular}{@{}l@{\hspace{3pt}}l@{\hspace{2pt}}p{0.66\linewidth}@{}}
{\scriptsize Q3-4B Base} & {\scriptsize \textcolor{lossred}{$\times$}} & {\scriptsize\raggedright \emph{``\ldots{} positioned closer to the camera, as it is in the foreground and its front is directly facing the viewer.''} $\rightarrow$ washing machine} \\
{\scriptsize Q3-4B Outcome} & {\scriptsize \textcolor{lossred}{$\times$}} & {\scriptsize\raggedright \emph{``\ldots{} the foreground of the image and occupies a more central position relative to the camera's perspective.''} $\rightarrow$ washing machine} \\
{\scriptsize Q3-4B Dense} & {\scriptsize \textcolor{lossred}{$\times$}} & {\scriptsize\raggedright \emph{``\ldots{} a more central position in the frame. The chair is further away, behind the table and to the left.''} $\rightarrow$ washing machine} \\
{\scriptsize \bfseries Q3-4B CALIPER} & {\scriptsize \textcolor{okgreen}{$\checkmark$}} & {\scriptsize\raggedright \emph{``\ldots{} and the washing machine is further back (in the mid-ground), the chair is closer to the camera.''} $\rightarrow$ \hla{rightmost chair}} \\
\noalign{\vspace{1pt}}
{\scriptsize IVL3-8B Base} & {\scriptsize \textcolor{okgreen}{$\checkmark$}} & {\scriptsize\raggedright \emph{``\ldots{} to the wall. - The \hlc{rightmost chair} is positioned closer to the foreground, nearer to the camera. Answer:''} $\rightarrow$ \hla{rightmost chair}} \\
{\scriptsize IVL3-8B Outcome} & {\scriptsize \textcolor{lossred}{$\times$}} & {\scriptsize\raggedright \emph{``\ldots{} front of the camera, while the \hlc{rightmost chair} is further away, placed at the end of the table. Answer:''} $\rightarrow$ washing machine} \\
{\scriptsize IVL3-8B Dense} & {\scriptsize \textcolor{lossred}{$\times$}} & {\scriptsize\raggedright \emph{``\ldots{} further back in the image compared to the \hlc{rightmost chair}, which is closer to the foreground. Answer:''} $\rightarrow$ washing machine} \\
{\scriptsize \bfseries IVL3-8B CALIPER} & {\scriptsize \textcolor{okgreen}{$\checkmark$}} & {\scriptsize\raggedright \emph{``\ldots{} to the wall. - The \hlc{rightmost chair} is positioned closer to the foreground, nearer to the camera. Answer:''} $\rightarrow$ \hla{rightmost chair}} \\
\end{tabular}\par\vspace{1pt}
{\scriptsize\itshape\color{caserule}Ground-truth answer: rightmost chair.}
\end{minipage}
\end{casebox}
\end{figure}

\begin{figure}[tbh]
\begin{casebox}{RoboSpatial-Home, both backbones on one item}
\noindent
\begin{minipage}[t]{0.20\linewidth}\vspace{0pt}\centering
\includegraphics[width=\linewidth]{figures/ac_both_robospatial.jpg}
\end{minipage}\hfill
\begin{minipage}[t]{0.76\linewidth}\vspace{0pt}
{\scriptsize Can the watch fit above the toilet? Answer yes or no.}\par\vspace{1.5pt}
\stage{stageone}{Perception}
{\scriptsize\emph{``The image shows a bathroom with a toilet on the left, a bathtub with a shower curtain in the center, and a wooden shelving unit on the right. - On the top\,\ldots{}''}}\par\vspace{2pt}
\stage{stagethree}{Inference}
\begin{tabular}{@{}l@{\hspace{3pt}}l@{\hspace{2pt}}p{0.66\linewidth}@{}}
{\scriptsize Q3-4B Base} & {\scriptsize \textcolor{lossred}{$\times$}} & {\scriptsize\raggedright \emph{``\ldots{} the watch cannot fit above the toilet as there is no such surface or space available in the image.''} $\rightarrow$ no} \\
{\scriptsize Q3-4B Outcome} & {\scriptsize \textcolor{lossred}{$\times$}} & {\scriptsize\raggedright \emph{``\ldots{} the shelf is not positioned to be directly above it. Therefore, the watch cannot fit above the toilet.''} $\rightarrow$ no} \\
{\scriptsize Q3-4B Dense} & {\scriptsize \textcolor{lossred}{$\times$}} & {\scriptsize\raggedright \emph{``\ldots{} above the toilet, and there is no indication that it could be moved there without obstruction.''} $\rightarrow$ no} \\
{\scriptsize \bfseries Q3-4B CALIPER} & {\scriptsize \textcolor{okgreen}{$\checkmark$}} & {\scriptsize\raggedright \emph{``\ldots{} and the watch is placed on that shelf. - Therefore, the watch is physically located above the toilet.''} $\rightarrow$ \hla{yes}} \\
\noalign{\vspace{1pt}}
{\scriptsize IVL3-8B Base} & {\scriptsize \textcolor{lossred}{$\times$}} & {\scriptsize\raggedright \emph{``\ldots{} a watch could be placed without interference from the shower curtain or the bathtub fixtures. Answer:''} $\rightarrow$ no} \\
{\scriptsize IVL3-8B Outcome} & {\scriptsize \textcolor{lossred}{$\times$}} & {\scriptsize\raggedright \emph{``\ldots{} above the toilet where a watch could be placed without interference from the shower curtain. Answer:''} $\rightarrow$ no} \\
{\scriptsize IVL3-8B Dense} & {\scriptsize \textcolor{okgreen}{$\checkmark$}} & {\scriptsize\raggedright \emph{``\ldots{} issues. Based on the observations, the watch is already placed above the toilet on the shelf. Answer:''} $\rightarrow$ \hla{yes}} \\
{\scriptsize \bfseries IVL3-8B CALIPER} & {\scriptsize \textcolor{okgreen}{$\checkmark$}} & {\scriptsize\raggedright \emph{``\ldots{} issues. Based on the observations, the watch is already placed above the toilet on the shelf. Answer:''} $\rightarrow$ \hla{yes}} \\
\end{tabular}\par\vspace{1pt}
{\scriptsize\itshape\color{caserule}Ground-truth answer: Yes.}
\end{minipage}
\end{casebox}
\end{figure}

\end{document}